 \documentclass[10pt,final,twocolumn]{IEEEtran}

\usepackage{xcolor}

\ifCLASSOPTIONcompsoc
\usepackage[nocompress]{cite}
\else
\usepackage{cite}
\fi
\usepackage{caption}

\ifCLASSINFOpdf
\usepackage[pdftex]{graphicx}
\graphicspath{{./fig/}}
\usepackage{epstopdf}
\DeclareGraphicsExtensions{.pdf,.jpeg,.png,.eps}
\else
\usepackage[dvips]{graphicx}
\graphicspath{{./fig/}}
\DeclareGraphicsExtensions{.eps}
\fi

\usepackage[cmex10]{amsmath}
\usepackage{amsmath,amsthm,amssymb,amsfonts,mathrsfs,bm}
\usepackage{algorithm,algorithmicx,algpseudocode,hyperref}
\usepackage{ragged2e}
\usepackage{tikz}
\usepackage[inline]{enumitem}
\usepackage{epsfig,bbm}

\usepackage{setspace}
\usepackage[export]{adjustbox}

\ifCLASSOPTIONcompsoc
\usepackage[caption=false,font=footnotesize,labelfont=sf,
textfont=sf]{subfig} 
\else
\usepackage[caption=false,font=footnotesize]{subfig}
\fi

\def \bS {{\bf S}}
\def \bx {{\bf x}}
\def \bY {{\bf Y}}

\def \bPhi {{\boldsymbol \Phi}}
\def \balpha {{\boldsymbol \alpha}}
\def \bphi {{\boldsymbol \phi}}
\def \bX {{\bf X}}
\def \bE {{\bf E}}
\def \bP {{\bf P}}
\def \bZ {{\bf Z}}
\def \bD {{\bf D}}
\def \bS {{\bf S}}
\def \bw {{\bf w}}
\def \ba {{\bf a}}
\def \bL {{\bf L}}
\def \bz {{\bf z}}

\def \bH {{\bf H}}
\def \by {{\bf y}}
\def \bA {{\bf A}}
\def \ba {{\bf a}}
\def \bQ {{\bf Q}}
\def \bq {{\bf q}}
\def \bK {{\bf K}}
\def \bG {{\bf G}}
\def \bk {{\bf k}}

\def \balpha {{\boldsymbol \alpha}}

\def \bbeta {{\boldsymbol \beta}}

\begin{document}
	

\title{Large-scale Kernel-based Feature Extraction \\ 
via Low-rank  Subspace Tracking on a Budget
}

\author{Fatemeh~Sheikholeslami,~\IEEEmembership{Student
    Member,~IEEE,}
Dimitris~Berberidis,~\IEEEmembership{Student~Member,~IEEE,}
  and~Georgios~B.~Giannakis,~\IEEEmembership{Fellow,~IEEE}
  \IEEEcompsocitemizethanks{\IEEEcompsocthanksitem
  	Part of this work was presented in IEEE International Workshop on Computational Advances in Multi-Sensor Adaptive Processing (CAMSAP), Curacao, Dutch Antilles, Dec. 2017 \cite{camsap}.\protect\\
  	This work was supported by NSF grants 1500713, 1514056, and the NIH grant no.  1R01GM104975-01. Preliminary parts of this work were presented at the {Proc. of Globalsip Conf.}, Orlando, FL, Dec. 2015.\protect\\
  	 Authors are with the Dept.\ of Electrical and Comp. Engr. and the
    Digital Tech. Center, University of Minnesota,
    Minneapolis, MN 55455, USA.\protect\\ 
    E-mails: \{sheik081,bermp001,georgios\}@umn.edu}
}

\markboth{IEEE Transactions on Signal Processing,~Vol.~XX, No.~XX, 2017. (submitted 5/2017) }%
{Sheikholeslami \MakeLowercase{\textit{et al.}}: Large-scale kernel-based feature extraction via budgeted nonlinear subspace-tracking }

\maketitle
\begin{abstract} 
Kernel-based methods enjoy powerful generalization capabilities in learning a variety of pattern recognition tasks. When such methods are provided with sufficient training data, broadly-applicable classes of nonlinear functions can be approximated  with  desired accuracy. Nevertheless, inherent to the {\em nonparametric} nature of kernel-based estimators are computational and memory requirements that become prohibitive with  large-scale datasets. In response to this formidable challenge, the present work puts forward a \emph{low-rank, kernel-based, feature extraction} approach that is particularly tailored for online operation. A novel generative model is introduced to approximate  high-dimensional (possibly infinite) features via a low-rank nonlinear subspace, the learning of which lends itself to a kernel function approximation. Offline and online solvers are developed  for the subspace learning task, along with affordable versions, in which the number of stored data vectors is confined to a  predefined budget. Analytical results provide performance bounds  on how well the kernel matrix as well as  kernel-based classification and regression tasks can be approximated by leveraging budgeted online subspace learning and feature extraction schemes. Tests on synthetic and real datasets demonstrate and benchmark the efficiency of the proposed method for dynamic nonlinear subspace tracking as well as online  classification and regressions tasks.
\end{abstract}


\IEEEpeerreviewmaketitle

\ifCLASSOPTIONcompsoc
\IEEEraisesectionheading{\section{Introduction}}
\else
\section{Introduction}
\fi

\IEEEPARstart{K}{ernel-based} expansions can boost the generalization capability of learning tasks by powerfully modeling nonlinear functions, when linear functions fall short in practice. When provided with sufficient training data, kernel methods can approximate arbitrary nonlinear functions with  desired accuracy. Although  ``data deluge'' sets the stage  by providing the ``data-hungry'' kernel methods with huge datasets, limited memory and computational  constraints prevent such tools from fully  exploiting their learning capabilities. In particular, given  $N$ training $D \times 1$ vectors $\{\bx_\nu\}_{\nu=1}^N$, kernel regression or classification machines take $\mathcal{O}(N^2D)$ operations to form the  $N \times N$ kernel matrix $\bK$, memory  $\mathcal{O}(N^2)$ to store it, and $\mathcal{O}(N^3)$ computational complexity to find the sought predictor or classifier.

In this context, several  efforts have been made in different fields of stochastic optimization, functional analysis, and numerical linear algebra  to speed up kernel machines for ``big data'' applications \cite{kivinen,Doublystochastic, pegasus,Drineas,Nystrom,Recht,lu}.
A common approach to scaling up  kernel methods is to approximate the kernel matrix $\bK$ by a low-rank factorization; that is,  $\bK \simeq \hat{\bK}:=\bZ^\top \bZ$, where $\bZ \in \mathbb{R}^{r \times N}$ with $r \;(\ll N)$ is the reduced rank, through which storage and computational requirements go down to $\mathcal{O}(Nr)$ and $\mathcal{O}(Nr^2)$, respectively. Kernel (K)PCA \cite{kernels} provides a viable factorization  for a such low-rank approximation, at the cost of  order  $\mathcal{O}(N^2r)$ computations. Alternatively, a low-rank factorization can be effected by randomly selecting  $r$ training vectors to approximate the kernel matrix \cite{RSVM}. 
Along these lines, Nystrom approximation \cite{Nystrom}, and its advanced renditions \cite{Drineas, impnyst,kumar,meka,ssnyst} are popular among this class of randomized factorizations. They trade off accuracy in approximating $\bK$ with $\hat{\bK}$, for reducing KPCA complexity from $\mathcal{O}(N^2r)$ to $\mathcal{O}(Nr)$. Their merits are well-documented for nonlinear regression and classification tasks performed offline \cite{cortes,mahdavi,sharp}. Rather than factorizing $\bK$, one can start from high-dimensional (lifted) feature vectors $\boldsymbol{\phi}(\bx_\nu)$ whose inner product induces the kernel $\kappa(\bx_i,\bx_j) := \langle \bphi(\bx_i),\bphi(\bx_j)\rangle$ \cite{Recht ,LLSVM, david,lu,allerton}. Approximating $\bphi(\bx)$ through an $r \times 1$ vector $\bz$, the nonlinear kernel can be approximated by a linear one as $\kappa(\bx_i,\bx_j) \simeq \bz_i^\top \bz_j$. Exploiting  the  fast linear learning machines \cite{liblinear, pegasus}, the kernel-based task then reduces to  learning a linear function over features $\{\bz_\nu\}_{\nu=1}^N$, which can be achieved in $\mathcal{O}(Nr)$ operations. Such a computationally attractive attribute is common to both kernel matrix factorization and lifted feature approximation. Note however, that online Nystrom-type schemes are \emph{not} available, while feature approximation algorithms are randomized, and thus they are \emph{not} data driven.

Different from kernel matrix and feature approximations performed in batch form, online kernel-based learning algorithms are of paramount importance. Instead of loading the entire datasets in  memory, online methods iteratively pass over the set from an external memory  \cite{kivinen, pegasus, CVM,LASVM,Paul,FastKPCA,ItKPCA}. This is also critical when the entire dataset is not available beforehand, but is acquired one datum at a time. For large  data streams however, as the number of  data increases with time, the support vectors (SVs) through which the function is estimated, namely the set $\mathcal{S}$ in the approximation $f(\bx) \simeq \hat{f}(\bx) = \sum_{i\in \mathcal{S}} \alpha_i \kappa(\bx_i,\bx)$, also increases in size. Thus, the function evaluation delay as well as the required memory for storing the SV set eventually become unaffordable. Efforts have been devoted to reducing the number of SVs while trying to maintain performance on unseen data (a.k.a. generalization capability) \cite{BSVM}. In more recent attempts, by restricting the maximum  number of SVs to a predefined {\em budget} $B$, the growth of algorithmic complexity  is confined to an affordable limit, that is maintained throughout the online classification \cite{twin, BSGD,forgetron} or regression \cite{BKLS} task. 

The present work builds  a  generative  model  according to which the high (possibly infinite)-dimensional features are approximated by their projection onto a  low-rank  subspace, thus providing  a linear kernel function approximation (Section~\ref{sec:pstatement}). In contrast to \cite{Recht, Drineas, impnyst, david}, where due to the nature of randomization the number of required features for providing an accurate  kernel function approximation  is often large,  systematically  learning the ambient nonlinear subspace yields an accurate approximation through a smaller number of extracted features.

 Offline and online solvers for  subspace learning  are developed, and their convergence is analyzed  in Sections~\ref{sec:offline} and~\ref{sec:OKFE} respectively.   In order to keep the complexity and memory requirements affordable, {\em budgeted} versions of the proposed algorithms are devised in Section~\ref{sec:OKFEB}, in which the number of stored data vectors is confined to a  predefined budget $B$. Budget maintenance is  performed through a greedy approach, whose effectiveness is corroborated through simulated tests. This is the first work to address {\emph{dynamic}} nonlinear (kernel-based) feature extraction under limited memory resources. 
 
 Analytical results in Section~\ref{sec:stability} provide performance bounds  on how well the kernel matrix as well as  kernel-based classification and regression  can be approximated by leveraging the novel budgeted online subspace-learning and feature-extraction approach.  Finally, Section~\ref{sec:tests} presents experiments on synthetic and real datasets, demonstrating the efficiency of the proposed methods in terms of accuracy and run time.

\section{Preliminaries and Problem Statement}\label{sec:pstatement}
Consider  $N$ real data vectors $\{\bx_\nu\}_{\nu=1}^N$ of size $D \times 1$. As large values of $D$ and $N$ hinder storage and processing of such datasets, extracting informative features from the data (a.k.a. {\it{dimensionality reduction}}) results in  huge savings on memory and computational requirements.  This fundamentally builds on the premise that the  informative part of the data is of low dimension $r<D$, and thus the data $\{\bx_\nu\}_{\nu=1}^N$ are well represented by the generative model 
\begin{equation}\label{model} 
	\bx_\nu = \bL \bq_\nu + \mathbf{v}_\nu  \;,  \quad \nu = 1,\ldots,N
\end{equation}
where  the tall
$D\times r$ matrix $\bL$ has rank $r<D$;  vector $\bq_\nu$ is the $r \times 1$
projection of $\bx_\nu$ onto the column space of $\bL$; and $\mathbf{v}_\nu$ denotes zero-mean additive noise. 

Pursuit of the subspace $\bL$ and the low-dimensional features $\{\bq_\nu\}_{\nu=1}^N$ is possible using  a blind least-squares (LS) criterion regularized by a rank-promoting term using e.g., the nuclear norm of $ \hat{\mathbf{X}} = \bL\bQ_N$, where $\bQ_N:= [\bq_1,...,\bq_N]$ \cite{fazel}. Albeit convex,  nuclear-norm regularization is not attractive for sequential learning.   

 To facilitate reducing the computational complexity, it is henceforth assumed that	an upper bound on the rank of matrix $\hat{\bX}$ is given $\rho \geq \text{rank}(\hat{\bX})$. \footnote{In practice, the rank is controlled by tuning regularization parameter, as it can be made small enough for sufficiently large $\lambda$.} Thus, building on the work of \cite{morteza_nuclear} by  selecting $r \geq \rho$, and to 
arrive at a scalable subspace tracker,  here we surrogate the nuclear norm with the summation of the Frobenious-norms of $\bL$ and $\bQ_N$, which yields (cf. Prop. 1 in \cite{morteza_nuclear} for proof on equivalence)
\begin{equation}\label{linearSubspace}
\min_{\bL ,\{ \bq_\nu\}_{\nu=1}^N} {\dfrac{1}{2N} \sum_{\nu=1}^n {\| \bx_\nu - \bL \bq_\nu \|_2^2}+ \dfrac{\lambda}{2N} \Big(\|\bL\|_F^2
+\|\bQ_N\|_F^2 \Big)}
\end{equation}
where $\lambda$ controls the tradeoff between  LS fit and  rank regularization \cite{morteza_jstsp}. Principal component analysis (PCA) - the ``workhorse'' of dimensionality reduction- solves \eqref{linearSubspace} when the rank regularization is replaced with orthonormality constraints on  $\bL$. 
Undoubtedly, the accuracy   of any linear dimensionality reduction method is dictated by how well the model \eqref{model} fits a given dataset, which is related to how well the corresponding data covariance matrix can be approximated by a low-rank matrix \cite[p.~534]{Hastie08}.

In practice however, low-rank linear models often fail to accurately capture the  datasets. A means to deal with nonlinearities in pattern recognition tasks, is to first map vectors $\{\bx_\nu\}_{\nu=1}^N$ to a higher $\bar{D}$-dimensional space  using a function $\boldsymbol{\phi}: \mathbb{R}^D \rightarrow \mathbb{R}^{\bar{D}}$ (possibly with ${\bar D}=\infty$), and subsequently seek a linear function over the lifted data $\boldsymbol{\phi} (\bx)$. This map induces a so-termed kernel function $\kappa(\bx_i,\bx_j) = \boldsymbol{\phi}^\top (\bx_{i}) \boldsymbol{\phi}(\bx_{j})$. Selecting the kernel to have a closed-form expression circumvents the need to explicitly know  $\{\boldsymbol{\phi}(\bx_\nu)\}_{\nu=1}^N  $ - what is referred to as  the  ``kernel trick.'' Similarly, the  norm corresponding to the reproducing kernel Hilbert space (RKHS) is defined as $\|\mathbf{\phi}(\mathbf{x})\|_{\mathcal{H}}^2 := \langle \boldsymbol{\phi}(\bx),\boldsymbol{\phi}(\bx) \rangle= \kappa(\bx,\bx)$. Upon defining the $\bar{D} \times N$ matrix  $\boldsymbol{\Phi}_N:= [\boldsymbol{\phi}(\bx_1),..., \boldsymbol{\phi}(\bx_N)]$, the $N\times N$ kernel matrix related to the covariance of the lifted data is formed with $(i,j)$ entry $\kappa(\bx_i,\bx_j)$ as   $\bK(\bx_{1:N},\bx_{1:N}) = \boldsymbol{\Phi}_N^\top \boldsymbol{\Phi}_N$, where $\bx_{1:N}:= \text{vec}[\bx_1, \bx_2, ..., \bx_N]$.  Its computation and storage incurs complexity  $\mathcal{O}(N^2D)$ and $\mathcal{O}(N^2)$ respectively, which is often not affordable when $N \gg$ and/or $ D \gg$.

Fortunately, $\bK$ for large data sets in practice has approximately low rank. This fact is exploited in e.g., \cite{LLSVM,Drineas} and \cite{Nystrom} to approximate $\bK$ via a low-rank factorization, hence reducing the evaluation and memory requirements of offline kernel-based learning tasks from $\mathcal{O}(N^2)$ down to $\mathcal{O}(Nr)$. Here, we further build on this observation to deduce that the low-rank property of   $\bK = \boldsymbol{\Phi}^\top_N\boldsymbol{\Phi}_N$ implies that $\boldsymbol{\Phi}_N$ can also be  approximated by a low-rank matrix, thus motivating our pursuit of {\it{online low-rank factorization}} of $\boldsymbol{\Phi}_N$. To this end, instead of projecting  $\{\bx_\nu\}$s onto the columns of $\bL$ as in \eqref{linearSubspace}, we will project   $\{\boldsymbol{\phi}(\bx_\nu)\}$s on $\bar{\bL}\in \mathbb{R}^{\bar{D}\times r}$, whose columns span what we refer to as  ``virtual'' column subspace  since $\bar{D}$ can be infinite. Specifically, we consider [cf. \eqref{linearSubspace}]
\begin{equation}\label{virtual}
\min_{{\bar{\bL} ,\{ \bq_\nu\}_{\nu=1}^N}} \dfrac{1}{2N} \sum_{\nu=1}^N{\| \boldsymbol\phi(\bx_\nu) - \bar{\bL} \bq_\nu \|_{\mathcal{H}}^2}+ \dfrac{\lambda}{2N} \Big(\|\bar{\bL}\|_{HS}^2 +  \| \bQ_N\|_F^2 \Big)
\end{equation} 
where the  $\ell_2$-norm has been substituted by the $\mathcal{H}$-norm in the $\bar{D}$-dimensional Hilbert space. Similarly, let the Hilbert\textendash Schmidt operator be defined as  $\|\bar{\bL}\|_{HS} = \sqrt{\text{Tr}(\bar{\bL}^\top \bar{\bL} )} :={\sqrt{\sum_{c=1}^r \| \bar{\mathbf{l}}_c\|_\mathcal{H}^2}}$ with $\bar{\mathbf{l}}_c$ denoting the $c$-th column of $\bar{\bL}$. Note that for Euclidean spaces, the Hilbert-Schmidt norm reduces to the  Frobenious norm.

Observe also that similar to the linear model in \eqref{linearSubspace}, upon removing the regularization terms and adding the orthonormality constraints on the columns of $\bar{\bL}$, \eqref{virtual} reduces to that of KPCA (without centering) in primal domain \cite[p. 429]{kernels}. The present formulation in \eqref{virtual} however, enables us to develop sequential  learning algorithms, which will later be enhanced with a tracking capability for dynamic datasets.

For a fixed $\bQ_N$, the criterion in \eqref{virtual} is minimized by
\begin{equation}\label{LA}
{\bar \bL}_N=  \boldsymbol{\Phi}_N\bQ_N^\top
\Big(\bQ_N\bQ_N^\top+{\lambda}\mathbf{I}\Big)^{-1} := \boldsymbol{\Phi}_N\bA\;
\end{equation}
where the $N \times r$ factor $\bA$ can be viewed as ``morphing' the columns of $\boldsymbol\Phi_N$ to offer a flexible basis for the lifted data.
Substituting \eqref{LA} back into \eqref{virtual} and exploiting the kernel trick, we arrive at
\begin{align}\label{L_kernel}
\min_{\bA ,\{ \bq_\nu \}_{\nu=1}^N} &\quad \dfrac{1}{2N} \sum_{\nu=1}^N\Big(\kappa(\bx_\nu,\bx_\nu)-2 \bk^\top(\bx_{1:N},\bx_\nu) \bA \bq_\nu \\& \qquad  \qquad + \bq_\nu^\top \bA^\top \bK(\bx_{1:N},\bx_{1:N}) \bA \bq_\nu  \Big) \nonumber\\ 
& + \dfrac{\lambda}{2N} \Big(\text{tr}\{\bA^\top \mathbf{K}(\bx_{1:N},\bx_{1:N}) \bA\} + \sum_{\nu=1}^N\| \bq_\nu\|_2^2 \Big)\nonumber
\end{align}
where the  $N\times 1$ vector $\bk (\bx_{1:N},\bx_n)$ in \eqref{L_kernel} is the $n$-th column of $\bK(\bx_{1:N},\bx_{1:N})$, and since $\bA$ has size $N \times r$, the minimization in \eqref{L_kernel} does not depend on $\bar{D}$.

Our goal is to develop and analyze   batch as well as   online solvers for  \eqref{L_kernel}. By pre-specifying an affordable complexity for the online solver, we aim at  a low-complexity algorithm where   subspace learning and feature extraction can be performed on-the-fly for streaming applications. Furthermore, we will introduce a novel approach to extracting features on which the kernel-based learning tasks of complexity $\mathcal{O}(N^3)$ can be well approximated by linear counterparts of complexity $\mathcal{O}(rN)$, hence realizing great savings in memory and computation while maintaining performance. A remark is now in order.

\noindent {\bf Remark 1.}  The subspace $\bar{\bL}_N$ in \eqref{LA} can be thought as a dictionary whose atoms are morphed via factor $\bA$. Sparse representation over kernel-based dictionaries have been considered  \cite{doublesparse,samplecomplexity,samplecomplexity2,k-coding}. Different from these approaches however, the novelty here is on developing algorithms that can process streaming datasets, possibly with dynamic underlying generative models. Thus, our goal is to efficiently learn and track a dictionary that adequately captures streaming data vectors,  and can afford a low-rank approximation of the underlying high-dimensional map.

\section{Offline kernel based Feature Extraction}\label{sec:offline}

Given a dataset $\{\bx_\nu\}_{\nu=1}^N$ and leveraging the bi-convexity of the  minimization in \eqref{L_kernel}, we introduce in this section a batch solver, where two blocks of variables ($\bA$ and $\{\bq_\nu\}_{\nu=1}^N$) are updated alternately. The following two updates are carried out iteratively until convergence.

{\bf{Update 1}}. With $\bA[k]$  given from iteration $k$, the projection vectors $\{\bq_\nu \}_{\nu=1}^N$ in iteration $k+1$ are updated as
\begin{subequations}
\begin{align}\label{q_t_off}
\bq_\nu[k+1] = \arg\min_{\bq}  \quad  & \ell(\bx_\nu;\bA[k],\bq;\bx_{1:N})+\dfrac{\lambda}{2} \| \bq \|_2^2
\end{align}
where the fitting cost $\ell(.)$  is given by [cf. \eqref{virtual}-\eqref{L_kernel}]
\begin{align}\label{ell}
\hspace*{-0.3cm}\ell(\bx_\nu;\bA[k],\bq;\bx_{1:N}) & :=  \dfrac{1}{2}\|\bphi(\bx_\nu) - \bPhi_N\bA[k] \bq\|_\mathcal{H}^2 \\ 
 &\; =  \kappa(\bx_\nu,\bx_\nu) \nonumber-2 \bk^\top(\bx_{1:N},\bx_\nu) \bA[k] \bq   \\
  &  \,\quad+ \bq^\top \bA^\top[k] \bK(\bx_{1:N},\bx_{1:N}) \bA[k]  \bq  \nonumber \; .
\end{align}
\end{subequations}
The minimizer of \eqref{q_t_off} yields the  features as regularized projection coefficients of the lifted data vectors onto the virtual  subspace $\bar{\bL}_N[k] = \boldsymbol{\Phi}_N\bA[k]$, and is given  in closed form by
\begin{align}\label{sol_q_off}
\bq_\nu[k+1] &=  (\bA^\top[k]\mathbf{K}(\bx_{1:N},\bx_{1:N})\bA[k]+\lambda \mathbf{I}_r) ^{-1} \nonumber \\
&\; \times  \bA^\top[k] \bk(\bx_{1:N},\bx_\nu), \quad \nu = 1,...,N\; .
\end{align}

{\bf{Update 2}}. With  $\{\bq_\nu[k+1] \}_{\nu=1}^N$ fixed and after dropping irrelevant terms, the subspace factor is obtained as [cf. \eqref{L_kernel}] 
\begin{align}\label{A_t}
\bA[k+1] =\arg \min_\bA & \dfrac{1}{N}\sum_{\nu=1}^N\ell(\bx_\nu; \bA,\bq_\nu[k+1];\bx_{1:N}) \nonumber \\
& + \dfrac{\lambda}{2N} \text{tr} \{ \bA^\top \mathbf{K}(\bx_{1:N},\bx_{1:N}) \bA\} \; .
\end{align}
Since $\bK$ is  positive definite in practice,  \eqref{A_t} involves a strictly convex minimization. Equating the gradient to zero, yields the wanted subspace factor in closed form
\begin{equation}\label{A_off}
\bA[k+1]=  \bQ_N^\top[k+1]
\Big(\bQ_N[k+1] \bQ_N^\top[k+1]+{\lambda}\mathbf{I}_r\Big)^{-1}  \;.
\end{equation}
Algorithm 1 provides the pseudocode for the update rules \eqref{sol_q_off} and \eqref{A_off} of the batch solver, and the following proposition gives a guarantee on the convergence of the proposed solver to a local stationary point.

\noindent
{\bf{Proposition 1}}. \emph{For positive definite kernels and $\lambda > 0$, the sequence $\{\bA[k],\bQ_N[k]\}$ generated by  Algorithm 1 converges to a stationary point of the minimization in  \eqref{L_kernel}.}

\noindent{\emph{Proof}}: Since the minimizations in \eqref{q_t_off} and \eqref{A_t} are strictly convex with unique solutions, the result follows readily from \cite[p. 272]{Bertsekas}. $\hfill \blacksquare$

\begin{algorithm}[t]\label{alg:offline}
\caption{BKFE: Batch Kernel-based Feature Extraction}\label{Maint}
\begin{algorithmic}
\State {\textbf{Input} $\{\bx_\nu\}_{\nu=1}^N, \lambda$}
\State {Initialize} $\bA{[1]}$ at random 
\State \textbf{For} $k=1,\ldots$ do
\State {$ \quad \bS[k+1] = \Big(\bA^\top[k]\mathbf{K}(\bx_{1:N},\bx_{1:N})\bA[k]+\lambda \mathbf{I}_r\Big)^{-1} \bA^\top[k]$}
\vspace{-0.4cm}
\State \begin{align*}
\bQ[k+1] =  \bS[k+1] \bK(\bx_{1:N},\bx_{1:N}) \qquad \qquad \qquad \qquad 
\end{align*}
\vspace{-1cm}
\State \begin{equation*}
  \quad\bA[k+1]=  \bQ_N^\top[k+1]
\Big(\bQ_N[k+1] \bQ_N^\top[k+1]+{\lambda}\mathbf{I}_r\Big)^{-1}  
\end{equation*}

\State \textbf{Repeat Until Convergence}
\State \textbf{Return} $\bA{[k]},  \{\bq_{\nu}[k] \}_{\nu=1}^N$
\end{algorithmic}
\end{algorithm}

Since matrix inversions in   \eqref{sol_q_off} and \eqref{A_off} cost $\mathcal{O}(r^3)$, and $\bQ_N$ and $\bA$ have size $r \times N$ and $N \times r$, respectively, the per iteration cost is $\mathcal{O}(N^2r + N r^2 + r^3)$. 
Although the number of iterations needed in practice for Algorithm 1 to converge is effectively small, this per iteration complexity can be unaffordable for large datasets. In addition,  datasets are not always available offline, or due to their massive volume,  can not be uploaded into memory at once. To cope with these issues,  an online solver for \eqref{L_kernel} is developed next, where the updates are carried out by iteratively  passing over the dataset one datum at a time.

\section{Online kernel based feature extraction}\label{sec:OKFE}
This section deals with low-cost, on-the-fly updates of the `virtual' subspace $\bar\bL$, or equivalently its factor $\bA$ as well as the features  $\{\bq_\nu\}$ that are desirable to keep up with streaming data. For such online updates, stochastic gradient descent (SGD) has well-documented merits, especially for \emph{parametric} settings. 
 However,  upon processing $n$ data vectors, $\bA$ has size  $n \times r$, which obviously grows with $n$. Hence, as the size of $\bA$ increases with the number of data, the task of interest is a  \emph{nonparametric} one.
 Unfortunately, performance of SGD  on nonparametric learning such as the one at hand is an uncharted territory. Nevertheless, SGD can still be performed on the initial formulation \eqref{virtual}, where solving for the virtual $\bar\bL$ constitutes a parametric task, not dependent on $n$.

 Starting with an update for $\bar{\bL}$, an update  for $\bA$ will be derived first, as an alternative to those in \cite{pegasus,Doublystochastic}, and \cite{BSGD}.
Next, an SGD iteration for $\bA$ will be developed in subsection \ref{SGDNonparam}, while in subsection \ref{Link} a connection between the two update rules will be drawn, suggesting how SGD can be broadened to learning nonparametric models as well. 
\subsection{SGD on ``parametric''  subspace tracking}

Suppose that $\bx_n$ is acquired at time $n$, posing the overall joint subspace tracking and feature extraction problem as [cf. \eqref{virtual}]
\begin{equation}\label{onlineSubspace}
\min_{\bar{\bL} ,\{ \bq_\nu\}_{\nu=1}^n} \dfrac{1}{2n} \sum_{\nu=1}^n {\| \boldsymbol\phi(\bx_\nu) - \bar{\bL} \bq_\nu \|_\mathcal{H}^2}+ \dfrac{\lambda}{2n} \Big(\|\bar{\bL}\|_{HS}^2 +  \| \bQ_n\|_F^2 \Big) \; .
\end{equation}

Using an alternating minimization approach, we update features and the subspace per data vector as follows.

{\bf Update 1}. Fixing the  subspace estimate at its recent value $\bar\bL[n-1]:= \bPhi_{n-1} \bA[n-1]$ from time $n-1$, the projection vector of the new data vector $\bx_n$ is found as [cf. \eqref{q_t_off}]
\begin{subequations}
\begin{align}\label{q_n}
\bq[n]& = \arg\min_{\bq} { \ell(\bx_n;\bA[n-1],\bq;\bx_{1:n-1})+\dfrac{\lambda}{2} \| \bq \|_2^2}
\end{align}
which through the kernel trick readily yields
\begin{align}\label{sol_q}
\bq[n] = & (\bA^\top[n-1]\mathbf{K}(\bx_{1:n-1},\bx_{1:n-1})\bA[n-1]+\lambda \mathbf{I}_r) ^{-1} \nonumber \\ & \times \bA^\top[n-1] \bk(\bx_{1:n-1},\bx_n)\;.
\end{align}
\end{subequations}
Although \eqref{sol_q} can be done for all the previous features $\{\bq_\nu\}_{\nu=1}^{n-1}$ as well, it is skipped in practice to prevent exploding complexity. In the proposed algorithm, feature extraction is  performed only for the most recent data vector $\bx_n$.

{\bf Update 2}. Having obtained $\bq[n]$,  the subspace update is given by solving
\begin{equation}\label{L_update}
\min_{\bar{\bL}} \dfrac{1}{n} \sum_{\nu=1}^n \bar{\cal L}(\bx_\nu; \bar{\bL},\bq[\nu]) 
\end{equation}
where 
$\bar{\cal L}(\bx_\nu; \bar{\bL},\bq[\nu]) := {\dfrac{1}{2}\| \boldsymbol\phi(\bx_\nu) - \bar{\bL} \bq[\nu] \|_\mathcal{H}^2}+ \dfrac{\lambda}{2n} \|\bar{\bL}\|_{HS}^2 \;.$
Solving \eqref{L_update} as time evolves, becomes   increasingly complex, and eventually  unaffordable. If data $\{\bx_\nu\}_{\nu=1}^n$ satisfy the law of large numbers, then  \eqref{L_update} approximates  
$\min_{\bar\bL} \mathbb{E}[\bar{\mathcal{L}}(\bx_\nu; \bar\bL,\bq_\nu) ]$, where expectation is with respect to the  unknown probability distribution of the data. To reduce complexity of the minimization, one typically resorts to stochastic approximation solvers, where by dropping the expectation (or the sample averaging operator), the `virtual' subspace update is
\begin{equation}\label{LSGD}
\bar\bL[n] = \bar\bL[n-1] - \mu_{n,L} \bar{\bG}_n
\end{equation}
with $\mu_{n,L}$ denoting a preselected stepsize, and $\bar{\bG}_n$  the gradient of the $n$-th summand in \eqref{L_update} given by 
\begin{align}\label{G_bar}
\bar{\bG}_n & := \nabla_{\bar\bL}\bar{\cal L}(\bx_n; \bar\bL[n-1],\bq[n]) \nonumber \\
& \; = - \Big(\boldsymbol\phi(\bx_n) - \bar\bL[n-1] \bq[n] \Big) \bq^\top[n] + \dfrac{\lambda}{n} \bar\bL[n-1]\nonumber \\ 
& \;= \boldsymbol{\Phi}_n \begin{bmatrix}
\bA[n-1] \bq[n] \bq^\top[n] \\
 -\bq^\top[n] 
 \end{bmatrix}+\dfrac {\lambda}{n} \boldsymbol{\Phi}_n \begin{bmatrix}
 \bA[n-1]\\
 \mathbf{0}_{1\times r}
 \end{bmatrix}\;.
\end{align}
Because $\bar{\bL}[n]$ has size $\bar{D}\times r$ regardless of $n$, iteration \eqref{LSGD} is termed ``parametric''
Using \eqref{LA} to rewrite  $\bar\bL[n] = \boldsymbol{\Phi}_{n}\bA[n]$, and  substituting into \eqref{LSGD}, yields 
 \begin{align}\label{Lrule}
 \boldsymbol{\Phi}_n   \bA[n] = &\boldsymbol{\Phi}_n \begin{bmatrix}
 \bA[n-1] \\
 \mathbf{0}_{1 \times r }
 \end{bmatrix}  \nonumber\\
 & - \mu_{n,L} \boldsymbol{\Phi}_n\begin{bmatrix}
 \bA[n-1]\Big(\bq[n] \bq^\top[n] +\dfrac{\lambda}{n} \mathbf{I}_r\Big)\\
 -\bq^\top[n] 
 \end{bmatrix} 
 \end{align}
which suggests the following update rule for factor  $\bA$
 \begin{equation}\label{A_param}
 \bA[n] =  \begin{bmatrix}
 \bA[n-1] -\mu_{n,L}\bA[n-1] \Big(\bq[n] \bq^\top[n]  + \dfrac{\lambda}{n} \mathbf{I}_r\Big) \\
 \mu_{n,L} \bq^\top[n] 
 \end{bmatrix}\;.
 \end{equation}
Even though \eqref{A_param} is not the only iteration satisfying \eqref{Lrule}, it offers an efficient update of the factor $\bA$. The update steps for the proposed parametric tracker are summarized as Algorithm 2.
 Note that the multiplication and inversion in \eqref{A_off} are avoided. However, per data vector processed, the kernel matrix is expanded by one row and one column, while the subspace factor $\bA$ grows accordingly by one row. 
\begin{algorithm}[t]\label{alg:online1}
\caption{Online kernel-based feature extraction with parametric update rule }\label{Maint}
\begin{algorithmic}
\State {\textbf{Input} $\{\bx_\nu\}_{\nu=1}^n , \lambda$}
\State {\textbf {Initialize}} $\bA{[1]}=\mathbf{1}_{1 \times r}$ , {$\bK(\bx_1,\bx_1) = \kappa(\bx_1,\bx_1)$}
\State \textbf{For} $n=2,\ldots$ do
\vspace{-0.4cm}
\State {\begin{align*}
 \bq[n] = & \;(\bA^\top[n-1]\mathbf{K}(\bx_{1:n-1},\bx_{1:n-1})\bA[n-1]+\lambda \mathbf{I}_r) ^{-1} \nonumber \\ & \times \bA^\top[n-1] \bk(\bx_{1:n-1},\bx_n)\;
\end{align*}}
\State $\:\;\bK(\bx_{1:n},\bx_{1:n})=\begin{bmatrix}
\bK(\bx_{1:n-1},\bx_{1:n-1}) \quad \bk(\bx_{1:n-1},\bx_n) \\
\bk^\top(\bx_{1:n-1},\bx_n) \qquad \quad  \kappa(\bx_n,\bx_n)
\end{bmatrix}$
\State \begin{equation*}
  \bA[n] =  \begin{bmatrix}
 \bA[n-1] -\mu_{n,L}\bA[n-1] \Big(\bq[n] \bq^\top[n]  + \dfrac{\lambda}{n} \mathbf{I}_r\Big) \\
 \mu_{n,L} \bq^\top [n]
 \end{bmatrix}
 \end{equation*}

\State \textbf{Return} $\bA{[n]},  \{\bq[{\nu}] \}_{\nu=2}^n$
\end{algorithmic}
\end{algorithm}

\subsection{SGD for ``nonparametric'' subspace tracking} \label{SGDNonparam}
In this subsection, the feature extraction  rule in \eqref{sol_q} is retained, while the update rule \eqref{A_param} is replaced by directly  acquiring the SGD  direction along the gradient of the instantaneous objective term with respect to $\bA$. Since, in contrast to the fixed-size matrix $\bar{\bL}$, the number of parameters in  $\bA$ grows with $n$, we refer to the solver developed in this subsection  as a \emph{nonparametric} subspace tracker. Furthermore, the connection between the two solvers is drawn in subsection \ref{Link}, and convergence of the proposed algorithm is analyzed in subsection \ref{Convergence}. 

At time instance $n$,  subproblem  \eqref{L_update} can be expanded using the kernel trick as 
 \begin{equation}\label{L_t}
\min_{{\bA\in \mathbb{R}^{n \times r}}} \dfrac{1}{n}\sum_{\nu=1}^n  {\cal L} (\bx_\nu; \bA, \bq[\nu] ; \bx_{1:n})\}
\end{equation} 
where 
	\begin{align}\nonumber
	{\cal L}(\bx_\nu; \bA,\bq[\nu];\bx_{1:n}) :=  &\,\,\ell(\bx_\nu; \bA,\bq[\nu];\bx_{1:n})\\ &+ \dfrac{\lambda}{2n} \text{tr} \{ \bA^\top \mathbf{K}(\bx_{1:n},\bx_{1:n}) \bA\}\;
	\end{align}
	with $\ell(.)$  given by \eqref{ell}. Stochastic approximation solvers of \eqref{L_t} suggest the update 
\begin{subequations}
{\begin{equation}\label{Aupdate2}
\bA[n] = \begin{bmatrix} 
\bA[n-1] \\
\mathbf{0}_{r \times 1}^\top
\end{bmatrix} - \mu_{n,A} {\bG}_n
\end{equation}
where $\mu_{n,A}$ denotes the user-selected step size, and $\bG_n$ denotes the gradient of the $n$-th summand in \eqref{L_t} with respect to $\bA$ that is given by
\begin{align}\label{G_t}
\bG_n
 := &
\nabla_\bA {\cal L} (\bx_n; [\bA^\top[n-1],\mathbf{0}_{r \times 1}]^\top,\bq[n];\bx_{1:n}) \nonumber \\
=&\, \bK(\bx_{1:n},\bx_{1:n}) \begin{bmatrix}
\bA[n-1] \\
\mathbf{0}_{r \times 1}^\top
\end{bmatrix}  \bq[n] \bq^\top[n]  \\
&
- \bk(\bx_{1:n},\bx_n) \bq^\top[n]+ \dfrac{ \lambda}{n}\bK(\bx_{1:n},\bx_{1:n})  \begin{bmatrix}
\bA[n-1] \\
\mathbf{0}_{r \times 1}^\top 
\end{bmatrix} \nonumber \;. 
\end{align}}
\end{subequations}
Substituting \eqref{G_t} into \eqref{Aupdate2} yields the desired update of $\bA$ which together with \eqref{sol_q} constitute our nonparametroc solver, tabulated under Algorithm 3.

\begin{algorithm}[t]\label{alg:online2}
\caption{Online kernel-based feature extraction with nonparametric update rule}\label{Maint}
\begin{algorithmic}
\State {\textbf{Input} $\{\bx_\nu\}_{\nu=1}^n , \lambda$}
\State {\textbf {Initialize}} $\bA{[1]}=\mathbf{1}_{1 \times r}$ , {$\bK(\bx_1,\bx_1) = \kappa(\bx_1,\bx_1)$}
\State \textbf{For} $n=2,\ldots$ do
\vspace{-0.4cm}
\State {\begin{align*}
 \bq[n] = & \;(\bA^\top[n-1]\mathbf{K}(\bx_{1:n-1},\bx_{1:n-1})\bA[n-1]+\lambda \mathbf{I}_r) ^{-1} \nonumber \\ & \times \bA^\top[n-1] \bk(\bx_{1:n-1},\bx_n)\;
\end{align*}}
\State $\bK(\bx_{1:n},\bx_{1:n})=\begin{bmatrix}
\bK(\bx_{1:n-1},\bx_{1:n-1}) \quad \bk(\bx_{1:n-1},\bx_n) \\
\bk^\top(\bx_{1:n-1},\bx_n) \qquad \quad  \kappa(\bx_n,\bx_n)
\end{bmatrix}$
\State {\begin{align}
\bG_n =&\,
\bK(\bx_{1:n},\bx_{1:n}) \begin{bmatrix}
\bA[n-1] \\
\mathbf{0}_{r \times 1}^\top
\end{bmatrix}  \bq[n] \bq^\top[n] \nonumber \\
&
- \bk(\bx_{1:n},\bx_n) \bq^\top[n]+ \dfrac{ \lambda}{n}\bK(\bx_{1:n},\bx_{1:n})  \begin{bmatrix}
\bA[n-1] \\
\mathbf{0}_{r \times 1}^\top 
\end{bmatrix} \nonumber \;
\end{align}}
\State \begin{equation*}
\bA[n] = \begin{bmatrix}
\bA[n-1] \\
\mathbf{0}_{r \times 1}^\top
\end{bmatrix} - \mu_{n,A} {\bG}_n
\end{equation*}

\State \textbf{Return} $\bA{[n]},  \{\bq[{\nu}] \}_{\nu=2}^n$
\end{algorithmic}
\end{algorithm}
\subsection{Parametric vis-a-vis  nonparametric SGD updates}\label{Link} 
Considering that  $\bar{\bL}[n] = \boldsymbol{\Phi}_n \bA[n]$ holds for all $n$, it is apparent from \eqref{G_t} and \eqref{G_bar} that $\bG_n = \boldsymbol{\Phi}_n^\top \bar{\bG}_n$.
The latter implies that  the update rule in \eqref{Aupdate2} amounts  to performing SGD on $\bar{\bL}$ with a matrix stepsize $\bD_n = \boldsymbol{\Phi}_n \boldsymbol{\Phi}_n^\top$; that is, 
\begin{equation}
\bar\bL[n] = \bar\bL[n-1] - \mu_{n,A} \bD_n\bar\bG_n \;.
\end{equation}
It is important to check whether this $\bD_n$ constitutes a valid descent direction, which is guaranteed since 
\begin{equation}
\bar{\bG}_n^\top \bD_n \bar{\bG}_n= \bH_n^\top \bK^\top(\bx_{1:n}, \bx_{1:n})\bK(\bx_{1:n}, \bx_{1:n}) \bH_n \succcurlyeq \mathbf{0}
\end{equation}
where 
\[\bH_n :=  \begin{bmatrix}
\bA[n-1] (\bq_n \bq_n^\top  + \dfrac{\lambda}{n} \mathbf{I}_r)\\
{{-\bq_n^\top}} 
\end{bmatrix}\; . \]
For positive-definite e.g., Gaussian kernel matrices, we have  $\bar{\bG}_n^\top \bD_n \bar{\bG}_n \succ \mathbf{0}$, which  guarantees that $-\bD_n \bar{\bG}_n$ is a descent direction \cite[~p. 35]{Bertsekas}. 
Leveraging this link, Algorithm 3 will be shown next to enjoy the same convergence guarantee as that of Algorithm 2.

\noindent 
{\bf{Remark 2.}} Although the SGD solver in Algorithm 3 can be viewed as a special case of  Algorithm 2, developing the parametric SGD solver in Algorithm 2 will allow us to analyze convergence of the two algorithms in the ensuing subsections.
\subsection{Convergence analysis}\label{Convergence}
The cost in \eqref{onlineSubspace}	 can be written as 
\begin{equation}
{F}_n (\bar\bL ) := \dfrac{1}{n} \sum_{\nu=1}^n \min_{\bq} {f_\nu} (\bx_\nu; \bar\bL, \bq )
\end{equation}
with $
f_\nu(\bx_\nu,\bar\bL,\bq):={\bar{\cal L}} (\bx_\nu; \bar\bL, \bq ) +  ({\lambda}/{2})\|\bq\|_2^2$, and $\bar{\cal L}$ as in \eqref{L_update}. 
Thus, the minimization in \eqref{onlineSubspace} is equivalent to $\min_{\bar\bL} F_n(\bar\bL)$.
To ensure  convergence  of the proposed algorithms, the following assumptions are adopted.

\noindent{{\bf{(A1)}} \emph{ $\{\bx_\nu\}_{\nu=1}^n$ independent identically distributed; and}

\noindent{\bf(A2) }\emph{The sequence $\{\|\bar\bL[\nu]\|_{HS}\}_{\nu=1}^\infty$ is bounded.}

Data independence across time is standard when studying the performance of online algorithms \cite{morteza_jstsp}, while  boundedness of the  iterates   $\{\|\bar\bL[\nu]\|_{HS}\}_{\nu=1}^\infty$, corroborated by simulations, is a technical condition that simplifies  the analysis, and    in the present setting is provided due to the Frobenious-norm regularization. In fact, rewriting subspace update in Alg. 2 yields
	 \[
	\bar{\bL}[n] =   \bar{\bL}[n-1] \Big( \mathbf{I} - \mu_{n,L} (\bq[n] \bq^\top[n] +\dfrac{\lambda}{n} \mathbf{I}_r)\Big)  + \mu_{n,L} \bphi_n \bq^\top,
	\]
	which    consists of: i)  contraction of  the most recent subspace iterate; and, ii) an additive term. Thus, with proper selection of the diminishing step size $\mu_{n,L}$, A2 is likely to hold. 
 The following proposition provides convergence guarantee for  the proposed algorithm.

\noindent{\bf{Proposition 2}}.
\emph{~Under (A1)-(A2), if $\mu_{n,L}= {1}/{\bar{\gamma}_n}  $ with $\bar{\gamma}_n:=\sum_{\nu=1}^n \gamma_\nu$ and $\gamma_\nu \geq \|\nabla^2 {\bar{\cal L}} (\bx_\nu; \bar\bL, \bq[\nu] )\|_\mathcal{H} \quad \forall
 n$, then the subspace iterates  in \eqref{LSGD} satisfy $\lim_{n \rightarrow \infty} \nabla F_n(\bar\bL[n]) = \mathbf{0}$ almost surely;  that is, $\textrm{Pr}\{\lim\limits_{n \rightarrow \infty } \nabla _{\bar{\mathbf{L}}} {F_n} (\bar{\bL}[n]) =0\} =1$, thus the sequence  $\{\bar{\bL}[\nu]\}_{\nu=1}^\infty$ falls into the stationary point of \eqref{onlineSubspace}.}

\noindent 
\emph{Proof}: Proof is inspired by \cite{Mairal}, and a sketch of the required modifications  can be found in the Appendix.
 
So far, we have asserted  convergence of the SGD-based algorithm for the ``virtual'' $\bar\bL$ provided by Algorithm 2. A related convergence result for Algorithm 3 is guaranteed by the following argument.

\noindent{\bf{Proposition 3}}.
\emph{ Under (A1)-(A2) and for positive definite radial kernels, if $\mu_{n,A}={1}/{\bar{\xi}_n} $ with $\bar{\xi}_n:= \sum_{\nu=1}^n \xi_n$ and $\xi_n \geq n \gamma_n$, then the subspace iterates  in \eqref{Aupdate2} satisfy $\lim_{n \rightarrow \infty} \nabla C_n(\bar\bL[n]) = \mathbf{0}$ almost surely;  that is, $\textrm{Pr}\{\lim_{n \rightarrow \infty} \nabla C_n(\bar\bL[n]) = \mathbf{0} \}=1$ , and the subspace  iterates will converge to the stationary point of \eqref{onlineSubspace}.}

\noindent{\emph{Proof:}} 
The proof follows the steps in Proposition 2, with an extra step in the construction of the appropriate surrogate cost in Step 1. 
	In particular, using that $\forall n$ the optimal subspace is of the form $\bar{\bL}_n=\bPhi_n \bA$, the objective $\tilde{f}_\nu$ can be further majorized  over the subset of virtual subspaces  $\bar\bL = \boldsymbol{\Phi}_n \bA$, by 
	\begin{align*}
	&\check{f}_n(\bx_n;\bPhi_n,\bA,\bq[n])  :=  f_n(\bx_n;\bar\bL[n-1],\bq[n]) \\
	& \qquad \quad +\text{tr}\{ \nabla_{\bar\bL} f_n(\bx_n;\bar{\bL}[n-1],\bq[n]) 		(\bPhi_n \bA-\bar\bL[n-1])^\top\}\nonumber \\
	&\qquad \quad+ \dfrac {\xi_n}{2} \| \bA -\begin{bmatrix}
	\bA[n-1]\\
	\mathbf{0}_{1 \times r}
	\end{bmatrix}\|_F^2 \nonumber
	\end{align*}
	for which we have
	\begin{align*}
	 \tilde{f}_n (\bx_n;\bar\bL,\bq[n])&-  \check{f}_\nu(\bx_\nu;\bPhi_n,\bA,\bq_\nu) \\
	& \hspace{-1.5cm}= \dfrac {\gamma_n}{2} \|\bar\bL-\bar\bL[n-1]\|_{HS}^2 - \dfrac {\xi_n}{2} \| \bA -\begin{bmatrix}
	\bA[n-1]\\
	\mathbf{0}_{1 \times r}
	\end{bmatrix}\|_F^2 \;.
	\end{align*}
The Cauchy-Schwarz inequality implies that
\begin{align*}
\|\bar\bL-\bar\bL[n-1]\|_{HS}^2 & = \|\bPhi_n \bA-\bPhi_n\begin{bmatrix}
	\bA[n-1]\\
	\mathbf{0}_{1 \times r}
	\end{bmatrix}\|_{HS}^2 \\
	& \leq \| \bPhi_n\|_{HS}^2  \| \bA-\begin{bmatrix}
	\bA[n-1]\\
	\mathbf{0}_{1 \times r}
	\end{bmatrix}\|_F^2
\end{align*}
	and by choosing $\xi_n \geq \| \bPhi_n\|_F^2\gamma_n =n \gamma_n $, we will have $ \tilde{f}_n (\bx_n;\bar\bL,\bq[n])\leq  \check{f}_\nu(\bx_\nu;\bPhi_n,\bA,\bq_\nu) $. Selecting now  $\check{f}_\nu(.)$ as the new surrogate whose minimizer coincides with the update rule in \eqref{Aupdate2}, the rest of the proof follows that of Prop. 2.  $\hfill\blacksquare$

\section{Reduced-complexity OK-FE  on a budget}\label{sec:OKFEB}

Per data vector processed, the iterative solvers of the previous section have one column of $\bPhi_n$ and one row of $\bA$ added, which implies growing memory and complexity requirements as  $n$ grows. The present section combines  two means of coping with this formidable challenge: one based on  {\em censoring} uninformative data, and the second based on {\emph{budget maintenance}}.
By modifying Algorithms 2 and 3 accordingly, memory and complexity requirements are rendered affordable.
\subsection{Censoring uninformative data}\label{sec:censoring}

In the LS cost that Algorithms 2 and 3 rely on,  small values of the fitting error can be tolerated in practice without noticeable performance degradation. This  suggests modifying the LS cost so that small  fitting errors (say up to $\pm \epsilon$) induce no penalty, e.g., by invoking the $\epsilon-$insensitive cost that is popular in support vector regression (SVR) settings \cite{Hastie08}.

Consider henceforth positive-definite kernels for which low-rank factors offer an approximation to the full-rank kernel matrix, and lead to a generally nonzero LS-fit   $\|\boldsymbol\Phi_n - \bar\bL \bQ_n\|_{\mathcal{H}}^2$. These considerations suggest replacing the LS cost $\ell(\bx_n;\bA[n-1],\bq;\bx_{1:{n-1}})$ with
\begin{align}\label{e-ins}
& \check{\ell}(\bx_n;\bA[n-1],\bq;\bx_{1:n-1}) \\
&    := \begin{cases} {0}
\qquad \qquad  \qquad \hspace{0.1cm}\mbox{if } {\ell}(\bx_n;\bA[n-1],\bq;\bx_{1:n-1})< \epsilon \\
 {\ell}(\bx_n;\bA[n-1],\bq;\bx_{1:n-1}) -\epsilon   \hspace{1.4cm}   \mbox{otherwise}. \end{cases} \nonumber
 \end{align}

By proper choice of $\epsilon$, the cost $\check{\ell}(.)$ implies that if  
${\ell}(\bx_n;\bA[n-1],\bq_n;\bx_{1:n-1})< \epsilon$, the virtual  
$\boldsymbol{\phi}(\bx_n)$ is captured well enough by the virtual current subspace $\bar{\bL}[n-1] = \boldsymbol{\Phi}_{n-1} \bA[n-1]$, and the solver will not attempt to  decrease its LS error, which suggests skipping the augmentation of $\bPhi_{n-1}$, provided by the new lifted datum $\bphi(\bx_n)$~\cite{censoring}. 

In short, if the upper branch of  \eqref{e-ins} is in effect, $\boldsymbol{\phi}(\bx_n)$ is deemed uninformative, and it is censored for the subspace update step; whereas having the lower branch deems $\boldsymbol{\phi}(\bx_n)$  informative, and augments the basis set of the virtual subspace. The latter case gives rise to what we term {\em online support vectors} (OSV), which must be stored, while `censored' data are discarded from subsequent subspace updates.

In order to keep track of the OSVs, let   $\mathcal{S}_{n-1}$  denote the set of indices corresponding to the SVs revealed up to time $n$. Accordingly, rewrite  $\bar\bL[n-1] = \bPhi_{\mathcal{S}_{n-1}} \bA[n-1]$, and the modified LS cost as ${\check{\ell}}(\bx_n;\bA[n-1],\bq;
\bx_{\mathcal{S}_{n-1}})$, depending on which of the following two cases emerges.

{\bf C1}. \emph{If  ${\check{\ell}}(\bx_n;\bA[n-1],\bq;
\bx_{\mathcal{S}_{n-1}})\leq \epsilon$, the OSV set will not grow, and we will have $\mathcal{S}_n = \mathcal{S}_{n-1}$; or,}

{\bf C2}. \emph{If  ${\check{\ell}}(\bx_n;\bA[n-1],\bq;
\bx_{\mathcal{S}_{n-1}})>\epsilon$, the OSV set will  grow, and we will have $\mathcal{S}_n = \mathcal{S}_{n-1} \cup \{n\}$.}

The subspace matrix per iteration will thus take the form $\bar{\bL}[n] = \bPhi_{\mathcal{S}_n} \bA[n]$, where $\bPhi_{\mathcal{S}_n} := [\boldsymbol\phi_{n_1},...,\boldsymbol\phi_{n_{|\mathcal{S}_n|}}]$, with $\mathcal{S}_n:=\{n_1,n_2,...,n_{|\mathcal{S}_n|}\} $, and $\bA \in \mathbb{R}^{|\mathcal{S}_n|\times r}$. 
Upon replacing $\bx_{1:n}$ in Algorithm 3 with $\bx_{\mathcal{S}_n}$, Algorithm 4 gives the pseudocode for our reduced-complexity online kernel-based feature extraction (OK-FE), which also includes a budget maintenance module that will be presented in the ensuing Section \ref{budget}.

Modifying the LS-fit in \eqref{e-ins} and discarding the censored data, certainly reduce the rate at which the  memory and complexity requirements increase.  In practice, thresholding is enforced after the budget is exceeded, when one needs to discard data. Regarding the selection of the threshold value,  the later may be initialized at zero and be gradually increased until the desired censoring rate is reached ( final threshold value will depend on the average fitting error and desired censoring rate) ; see also  \cite{censoring} for related issues.  Albeit at a slower rate, $|\mathcal{S}_n|$ may still grow unbounded as time proceeds. Thus, one is motivated to restrict the number of OSVs to a prescribed affordable budget, $|\mathcal{S}_n|\leq B$, and introduce a solver which maintains such a budget throughout the iterations. To this end, we introduce next a greedy `budget maintenance' scheme.

\subsection{Budget maintenance}\label{budget}

When inclusion of a new data vector into the OSV set pushes its cardinality $|\mathcal{S}_n|$ beyond the prescribed budget $B$, the budget maintenance module will discard one SV from the SV set. The removal strategy is decided according to a predefined rule. In the following, we will describe two strategies for budget maintenace. 

\subsubsection{Minimum-distortion removal rule} 
In this scheme, the  SV whose exclusion distorts the subspace   $\bar{\bL}[n]$ minimally will be discarded. Specifically, with $\bPhi_{n\setminus i}$ and $ \bA_{\setminus i}[n]$ denoting $\bPhi_n$ and $\bA[n]$ devoid of their $i$-th column and row, respectively, our rule for selecting the index to be excluded is
\begin{align}\label{Bud}
 i_* & = \arg\min _{i\in \mathcal{S}_n} \|\bPhi_n \bA[n] - \bPhi_{n\setminus i} \bA_{\setminus i}[n]\|_{HS}^2 \nonumber \\
 & =  \arg\min _{i\in \mathcal{S}_n} \text{tr}\{\bA^\top[n] \bK(\bx_{\mathcal{S}_n},\bx_{\mathcal{S}_n }) \bA[n] \\ & \hspace {1.9 cm}- 2 \bA_{\setminus i}^\top[n] \bK(\bx_{\mathcal{S}_n \setminus i},\bx_{\mathcal{S}_n }) \bA[n] \nonumber \\ & \hspace {1.9 cm} + \bA_{\setminus i}^\top[n] \bK(\bx_{\mathcal{S}_n \setminus i},\bx_{\mathcal{S}_n \setminus i}) \bA_{\setminus i}[n] \} \nonumber\;.
\end{align}
Enumeration over $\mathcal{S}_n$ and evaluation of the cost incurs complexity  $\mathcal{O}(B^3)$ for solving \eqref{Bud}.  Hence, in order to mitigate the computational complexity, a greedy scheme is put forth. Since exclusion of an SV will result in removing the corresponding row from the subspace factor, discarding the SV corresponding to the   row with the smallest $\ell_2-$norm suggests a reasonable heuristic greedy policy. To this end, one needs to find the index  
\begin{equation}\label{Bud2}
\hat{i}_* = \arg \min_{i=1,2,...,B+1} {\|\ba_i[n]\|_2}
\end{equation}
where $\ba^\top_i [n]$ denotes the $i-$th row of $\bA[n]$. Subsequently, $\hat{i}_*$ as well as the corresponding SV are discarded from  $\mathcal{S}_n$ and the SV set respectively, and an  OSV set of cardinality $|\mathcal{S}_n|=B$ is maintained.

\noindent {\bf Remark 3. }In principle,  methods related to those in {{\cite{BSGD}}}, including replacement of two SVs by a linear combination of the two, or projecting an SV on the SV set and discarding the projected SV, are also viable alternatives. In practice however, their improved performance relative to \eqref{Bud2} is negligible and along with their increased complexity, renders such alternatives less attractive for large-scale datasets.

\subsubsection{Recency-aware  removal  rule}
This policy is tailored  for tracking applications, where the subspace capturing the data vectors can change dynamically. As the subspace evolves, the fitting error will gradually increase, indicating the gap between the true and learned subspace, thus requiring incorporation of  new vectors into the subspace. In order for the algorithm to track a  dynamic subspace on a fixed budget, the budget maintenance module must gradually discard outdated SVs inherited from ``old'' subspaces, and include  new SVs. Therefore, apart from  ``goodness-of-fit'' (cf.~\eqref{Bud2}), any policy tailored to tracking should also take into account ``recency'' when deciding which SV is to be discarded.

 To this end, corresponding to the $i$-th SV, let  $\eta_i$ denote the recency factor whose value is initialized to 1. For every inclusion of a new SV, the recency $\eta_i$ of the current SVs will be degraded by a forgetting factor $0<\beta  \leq 1$; that is, $\eta_i$ will be replaced by $ \beta \eta_i$. Consequently, older SVs will have smaller $\eta_i$ value whereas  recent vectors will have $\eta_i \simeq 1$. To incorporate this memory factor into the budget maintenance module, our idea is to  choose the SV to be discarded according to 
 \begin{equation}\label{Bud3}
 \hat{i}_* = \arg \min_{i=1,2,...,B+1} {\eta_i \|\ba_i[n]\|_2}
 \end{equation}
which promotes discarding  older SVs over more recent ones.

By tuning $\beta$, the proposed memory-aware budget maintenance module can cover a range of different schemes. For large values of $\beta \eqsim 1$, it follows that $\eta_i\approx{\eta_j}~\forall{i,j\in{\mathcal{S}}}$, and \eqref{Bud3} approaches the minimum distortion removal rule in \eqref{Bud2}, which is tailored for learning static subspaces. On the other hand, for small $\beta$, the discarding rule is heavily biased towards removing old SVs rather than the newly-included ones, thus pushing the maintenance strategy towards a first-in-first-out (FIFO) approach, which is often optimal for  applications with fast-varying subspaces. 
Algorithms 4 and 5 tabulate the updates and the  greedy budget maintenance scheme, respectively. Budget maintenance strategy in \eqref{Bud2} is a special case of Alg. 5 with $\beta = 1$.

\begin{algorithm}
\caption{Online Kernel-based Feature Extraction on a Budget (OKFEB)}\label{Solver}
\begin{algorithmic}
\State  \textbf{Input} $\{\bx_\nu\}_{\nu=1}^n, \lambda $
\State {Initialize} $\bA{[1]}$ at random and $\mathcal{S}_1=\{1\}$
\State \textbf{For} $n=2,\ldots$ do
\vspace{-0.5cm}
\State \begin{align*}
\bq[n] = & (\bA^\top[n-1]\mathbf{K}(\bx_{\mathcal{S}_{n-1}},\bx_{\mathcal{S}_{n-1}})\bA[n-1]+\lambda \mathbf{I}_r)^{-1} \nonumber \\ & \times \bA^\top[n-1] \bk(\bx_{\mathcal{S}_{n-1}},\bx_n)
\end{align*}
\vspace{-1cm}
\State \begin{align*}
&  \ell_n =  k(\bx_n,\bx_n) -2 \bk^\top(\bx_{\mathcal{S}_{n-1}},\bx_n) \bA[n-1] \bq[n] \nonumber \\
&\qquad  + \bq^\top_n \bA^\top[n-1] \bK(\bx_{\mathcal{S}_{n-1}},\bx_{\mathcal{S}_{n-1}}) \bA[n-1]  \bq[n]
\end{align*}

\If {$\ell_n < \epsilon$ }
{ $\mathcal{S}_n = \mathcal{S}_{n-1}$}
\Else
\State $\mathcal{S}_n = \mathcal{S}_{n-1} \cup \{n\}$
\vspace{-0.5cm}
\State \begin{align*}
\check{\bG}_n
=&\,
\bK(\bx_{\mathcal{S}_{n}},\bx_{\mathcal{S}_{n}}) \begin{bmatrix}
\bA[n-1] \\
\mathbf{0}_{r \times 1}^\top
\end{bmatrix}  \bq[n] \bq^\top[n]\\
& - \bk(\bx_{\mathcal{S}_{n}},\bx_n) \bq^\top[n]
+ \dfrac{ \lambda}{n}\bK(\bx_{\mathcal{S}_{n}},\bx_{\mathcal{S}_n})  \begin{bmatrix}
\bA[n-1] \\
\mathbf{0}_{r \times 1}^\top
\end{bmatrix}
\end{align*}
\vspace{-0.2cm}
\State $\bA[n] = \begin{bmatrix}
\bA[n-1] \\
\mathbf{0}_{r \times 1}^\top
\end{bmatrix} - \mu_{n,A} \check{\bG}_n $
	\If {$|\mathcal{S}_n|>B$}
{ \text{Run budget maintenance module}}
\EndIf
\EndIf
\State \textbf{EndFor}
\State \textbf{Return} $\bA{[n]},\mathcal{S}_{n},  \{\bq[{\nu}] \}_{\nu=1}^n$
\end{algorithmic}
\end{algorithm}
\begin{algorithm}
\caption{Budget maintenace module}\label{Maint}
\begin{algorithmic}

\State {\textbf{Input} $\{\mathcal{S},\bA\, , \{\eta_i\}_{   i \in \mathcal{S}}\}$}
\State $\eta_i \leftarrow {\beta \eta_i  } \qquad  {  \forall i \in \mathcal{S}}$
\State $\hat{i}_* = \arg\min_{i\in \mathcal{S}} \eta_i \|\ba^\top_i\|_2$
	\State $\mathcal{S}\leftarrow \mathcal{S}\setminus \{\hat{i}_*\}$
	\State {Discard the $\hat{i}_*$-th row of $\bA$  and $\eta_{\hat{i}_*}$}
	\State {\textbf{Return} $\{\mathcal{S},\bA , \{\eta_i\}_{   i \in \mathcal{S}}\}$}
\end{algorithmic}
\end{algorithm}

\subsection{Complexity analysis}

 Computational complexity of the proposed OK-FEB  is evaluated in the present section. The computations required by the $n-$th iteration of Alg.~\ref{Solver} for feature extraction and parameter update depend on $B, r$, and $ D$, as well as the censoring process outlined in Section \ref{sec:censoring}. Specifically, computing $\check{\mathbf{G}}_n$ and performing the first-order stochastic update that yields $\mathbf{A}[n]$ requires $\mathcal{O}(B^2r)$ multiplications, a cost that is saved for skipped instances  when $\ell_n<\epsilon$. Regarding the computation of $\mathbf{q}[n]$, $Br(B+r)$ multiplications are needed to form $\bA^\top[n-1]\mathbf{K}(\bx_{\mathcal{S}_{n-1}},\bx_{\mathcal{S}_{n-1}})\bA[n-1]$, and $\mathcal{O}(r^3)$ multiplications for the inversion of $\bA^\top[n-1]\mathbf{K}(\bx_{\mathcal{S}_{n-1}},\bx_{\mathcal{S}_{n-1}})\bA[n-1]+\lambda \mathbf{I}_r$. Fortunately, the aforementioned computations can also be avoided for iteration $n$, if the previous iteration  performs no update on $\mathbf{A}[n-1]$; in this case, $(\bA^\top[n-1]\mathbf{K}(\bx_{\mathcal{S}_{n-1}},\bx_{\mathcal{S}_{n-1}})\bA[n-1]+\lambda \mathbf{I}_r)^{-1}$ remains unchanged and can simply be accessed from memory. Nevertheless, a ``baseline'' of computations is required for feature extraction related operations that take place regardless of censoring. Indeed, forming $\bA^\top[n-1] \bk(\bx_{\mathcal{S}_{n-1}},\bx_n)$ requires $Br$ multiplications for the matrix-vector product, and $\mathcal{O}(BD)$ for the evaluation of $B$ kernels in $\bk(\bx_{\mathcal{S}_{n-1}},\bx_n)$; the matrix-vector product that remains for obtaining $\mathbf{q}[n]$ requires $r^2$ additional multiplications. 
 
 Overall, running OK-FEB on $N$ data and with a value of $\epsilon$ such that $\check{N}\leq N$ data are used for updates requires $\mathcal{O}(\check{N}(Br(B+r)+r^3)+N(B(D+r)+r^2))$. Alternatively, tuning $\epsilon$ such that $\Pr\{\ell_n>\epsilon\}=\mathbb{E}[\check{N}/N]:=\rho$ yields an expected complexity $\mathcal{O}(N(Br(\rho(B+r)+1)+(\rho r +1)r^2 + BD))$.   As simulation tests will corroborate, the budget parameter $B$ can be chosen as $B = c r$ with $c \in [1.5 , 5]$. Thus, we can simplify the overall complexity order as $\mathcal{O}(Nr^2(\rho r +1) + NDr)$.

\section{Stability  of  kernel approximation}\label{sec:stability}

In this section, the effect of low-rank approximation of the lifted vectors on kernel-matrix approximation as well as kernel-based classification and regression is analytically quantified. 
Recall that given $\{\bx_\nu\}_{\nu=1}^N$, the virtual subspace obtained by running  OK-FEB    is $\bar{\bL} = \bPhi_{\mathcal{S}} \bA  \in \mathbb{R}^{\bar{D}\times r}$, and the corresponding  projection coefficients  are $\bQ_N$.  
By defining  the random variables $e_i := \|\bphi(\bx_i) - \hat{\bphi}(\bx_i)\|_\mathcal{H}^2 = \|\bphi(\bx_i)-\bar{\bL}\bq_i\|_\mathcal{H}^2$capturing the LS error, we have the following result.

\noindent{\bf{Proposition 4}}.  \emph {If  the random variables $e_i \in [0 \; , \; 1]$  are i.i.d.  with mean  $\bar{e} := \mathbb{E}[e_i]$, then for kernels satisfying $|\kappa{(\bx_i,\bx_j)}| \leq 1$, the matrix   $\bK =\bPhi^\top \bPhi$ can be approximated by $\hat{\bK} :=\hat{\bPhi}^\top \hat{\bPhi} $, and with  probability  at least $1-2e^{-2Nt^2}$, it holds that 
\begin{equation}
\dfrac{1}{N}\|\bK-\hat\bK\|_F \leq \sqrt{\bar{e}+t} \;(\sqrt{\bar{e}+t}+2)\;.
\end{equation}}
\noindent \emph{Proof:} Upon defining $\bar{\bE} := \hat{\bPhi} -\bPhi$, one can write 
\begin{subequations}
\begin{align}
\|\bK-\hat{\bK}\|_F&  =\|\bPhi^\top \bPhi - \hat{\bPhi}^\top \hat{\bPhi}\|_F  \nonumber\\
& = \|\bPhi^\top \bPhi - {(\bPhi+\bar{\bE})}^\top {(\bPhi+\bar{\bE})}\|_F \nonumber\\ & =\|\-2 \bar
{\bE}^\top \bPhi+ \bar{\bE}^\top \bar{\bE}\|_{F} \nonumber\\
& \leq 2 \|\bar{\bE}\|_{HS}\|\bPhi\|_{HS}+\|\bar{\bE}\|_{HS}^2 \label{uneq1}\\
& \leq 2\sqrt{N} \|\bar{\bE}\|_{HS} + \|\bar{\bE}\|_{HS}^2\label{uneq2}
\end{align}
\end{subequations}
where in \eqref{uneq1} we used the triangle inequality for the Frobenious norm along with the property $\|\mathbf{B}\mathbf{C}\|_{F} \leq \|\mathbf{B}\|_F\|\mathbf{C}\|_F$, and \eqref{uneq2} holds because for, e.g., radial kernels satisfying $|\kappa(\bx_i,\bx_j)|\leq 1$, we have 
\[\|\bPhi\|_{HS} := \sqrt{\text{tr}(\bPhi^\top \bPhi)}= 
 \sqrt{\sum_{i=1}^N\kappa(\bx_i,\bx_i)} \leq \sqrt{N}\;.  \]

 Furthermore, since $\|\bar{\bE}\|_F := \sqrt{\sum_{i=1}^N e_i}$, and $\bar{e}_N :=(1/N) \sum_{i=1}^N e_i$  with $e_i \in [0, 1]$, Hoeffding's inequality  yields
${\rm{Pr}}\Big(\bar{e}_N-\bar{e}\geq t\Big) \leq e^ {-2Nt^2}$,
which in turn implies
\begin{equation} 
{\rm{Pr}}\Big(\dfrac{1}{N}\|\bar{\bE}\|_F^2\geq \bar{e}+t \Big) = {\rm{Pr}}\Big(\bar{e}_N\geq \bar{e}+t \Big) \leq  e^ {-2Nt^2} \; .
\end{equation}
Finally, taking into account \eqref{uneq2}, it follows that with probability at least $1-2 e^ {-2Nt^2}$, we have 
\begin{equation}
\|\bK-\hat{\bK}\|_F \leq  N(2 \sqrt{\bar{e}+t} + (\bar{e}+t))\;\hfill .\blacksquare 
\end{equation}

 Proposition 4 essentially bounds the kernel approximation mismatch based on  how well the projection onto the subspace approximates the lifted data $\bphi(\bx)$.  

\noindent{\bf Remark 4}. Consider now decomposing the kernel matrix as 
\begin{align}
\hat{\bK}:=\hat{\bPhi}^\top \hat{\bPhi}=\,&(\bar{\bL} \bQ)^\top(\bar{\bL}\bQ)= \bQ^\top \bA^\top \bPhi_{\mathcal{S}}^\top \bPhi_\mathcal{S} \bA \bQ \nonumber \\
 =\,&\bQ^\top \bA^\top \bK_\mathcal{S} \bA \bQ =\bZ^\top \bZ
\end{align} 
where matrix  $ \bZ:= \bK_{\mathcal{S}}^{1/2} \bA \bQ$ has size $|\mathcal{S}| \times N$, and $\mathcal{S}$ denotes the budgeted SV set. 
This factorization of $\hat{\bK}$ could have resulted from a linear kernel over the $|\mathcal{S}| \times 1$ training data vectors forming the $N$ columns of  $\bZ$.  Thus, for kernel-based tasks such as kernel classification, regression, and clustering applied to large datasets, we can simply map the $D \times N$ data $\bX$ to the corresponding features $\bZ$ trained via the proposed solvers, and then simply rely on {\emph {fast linear}} learning methods to approximate the original kernel-based learning task; that is to approximate the function $f(\bx) = \sum_{i \in \mathcal{S}} c_i \kappa(\bx,\bx_i)$ by the linear function $g(\bz) = \bw^\top \bz$ expressed via the extracted features. Since  linear pattern recognition tasks incur complexity $\mathcal{O}(NB^2)$, they scale extremely well for large datasets (with $N \gg$), compared to kernel SVM that incurs complexity $\mathcal{O}(N^3)$. Furthermore, in the testing phase, evaluation of function  $f(\bx)$  requires  $\kappa(\bx_\nu,\bx_i)$ for $\forall i \in \mathcal{S}$ to be evaluated at complexity $\mathcal{O}(\mathcal{|S|} D )$, where $|\mathcal{S}|$ is the number of SVs that typically grows with $N$. In contrast, if approximated by the linear  $g(\bz)$, function evaluation requires $\mathcal{O}(BD + Br )$ operations including the feature extraction and function evaluation. Setting the budget  $B$ to 1.5 to 5 times the rank parameter $r$, our complexity is of order $\mathcal{O}(rD+r^2)$}, which represents a considerable  decrease over $\mathcal{O}(|\mathcal{S}|D)$.

Subsequently, we wish to quantify how the performance of linear classification and regression based on the features $\bK_\mathcal{S}^{1/2}\bA\bQ$  compares to the one obtained when training with the exact kernel matrix $\bK$. 

\subsection{Stability analysis for kernel-based classification}

Kernel-based  SVM classifiers  solve \cite[p.~205]{kernels}
\begin{align}\label{dual_full}
\balpha^* =\arg\min_{\balpha}&\; { \dfrac{1}{2}\balpha^\top \bY \bK \bY \balpha - \mathbf{1}^\top \balpha}\\ 
\text{s.t.}& \;\by^\top \balpha = 0 \nonumber \, \; \mathbf{0}\leq \balpha \leq \dfrac{C}{N}\mathbf{1}_N 
\end{align}
where $\bY$ is the diagonal matrix with the $i$-th label $y_i$ as its $i$-th diagonal entry, $\by^\top := [y_1,y_2,...,y_N]$, and $\mathbf{1}_N$ is an $n \times 1$ vector of $1$'s. Solution \eqref{dual_full} corresponds to the dual variables of the primal optimization problem, which yields 
\begin{align}\label{primal_full}
\bar{\bw}^* =\arg\min_{\bar{\bw}\in \mathbb{R}^{\bar{D}}}&\; {\dfrac{1}{2} \|\bar{\bw}\|_\mathcal{H}^2 + 
\dfrac{C}{N}\sum_{i=1}^N {\max \{0,1-y_i \bar{\bw}^\top \bphi(\bx_i)\}}}\; .
\end{align}
Here, parameter $C$ controls the trade-off between  maximization of the margin $1/\|\bw\|_\mathcal{H}$, and  minimization of the  misclassification penalty, 
while the solution of \eqref{primal_full} can be expressed as  $\bar{\bw}^* = \sum_{i=1}^N \alpha_i^* y_i \bphi(\bx_i)$\cite[~p.187]{kernels}.

Exploiting the reduced memory requirement offered through the low-rank approximation of the kernel matrix via OK-FEB, the dual problem can be approximated as 
\begin{align}\label{dual_apprx}
\hat{\balpha}^* =\arg\min_{{\balpha}}&\; {\dfrac{1}{2}\balpha^\top \bY\hat{\bK} \bY\balpha - \mathbf{1}^\top \balpha}\\ 
\text{s.t.}& \;\by^\top \balpha = 0 \nonumber \;  , \mathbf{0}\leq \balpha \leq \dfrac{C}{N}\mathbf{1}_N 
.
\end{align}
Viewing $\hat{\bK}$  as a linear kernel matrix over $\{\hat{\bphi}(\bx_i)\}$s (cf. Remark 4),  similar to \eqref{dual_full}, the minimization   \eqref{dual_apprx} can  be re-written in the  primal form  as
 \begin{align}\label{primal_apprx}
\hat{\bar{\bw}}^* =\arg\min_{\bar{\bw}}&\; {\dfrac{1}{2} \|\bar{\bw}\|_\mathcal{H}^2 + 
\dfrac{C}{N}\sum_{i=1}^N {\max \{0,1-y_i \bar{\bw}^\top \hat{\bphi}(\bx_i)\}}}
\end{align}
for which we have  $\hat{\bar{\bw}}^* = \sum_{i=1}^N \hat{\alpha}_i^* y_i \hat{\bphi}(\bx_i)$. 
Upon defining the random variable  $\xi_i :=  \|\bphi(\bx_i) -\hat{\bphi}(\bx_i)\|_\mathcal{H}$ with expected value  $\bar{\xi} := \mathbb{E}[\xi_i]$, the following proposition quantifies the gap between $\bar{\bw}^*$ and $\hat{\bar{\bw}}^*$.

\noindent{\bf{Proposition 5}}. \emph{If  $\xi_i \in [0 \; , \; 1]$ are i.i.d., with mean  $\bar{\xi}$, the mismatch between the linear classifiers given by \eqref{primal_full} and \eqref{primal_apprx}  can be bounded, and with probability at least $1-e^{-2Nt^2}$, we have 
\begin{equation}
\|\Delta \bw\|_{\mathcal{H}}^2 := \|{\bar{\bw}}^* - \hat{\bar{\bw}}^*\|_{\mathcal{H} }^2  \leq 2{C^{3/2}}  \Big( {\bar{\xi}+t} \Big) \;.
\end{equation}
}
\noindent \emph{Proof:} It clearly holds that 
\begin{align}
\|\Delta \bw\|_{\mathcal{H}}^2 & \leq \dfrac{C}{N}(\|\bar{\bw}^*\|_{\mathcal{H}}+\|\hat{\bar{\bw}}\|_{\mathcal{H}}){\sum_{i=1}^N { } \|\bphi(\bx_i)-\hat{\bphi}(\bx_i)\|_\mathcal{H}} \nonumber\\
& \leq  \dfrac{2 C^{3/2}  } {N}{\sum_{i=1}^N \|\bphi(\bx_i)-\hat{\bphi}(\bx_i)\|_\mathcal{H}} \leq 2 {C^{3/2}} ({\bar{\xi}}+t) \nonumber
 \end{align}
where the first inequality relies on the strong convexity of  \eqref{primal_full}, \eqref{primal_apprx}, and the fact that $\|\bar{\bw}^*\|_\mathcal{H} \leq \sqrt{C}$ and $\|\hat{\bar{\bw}}^*\|_\mathcal{H} \leq  \sqrt{C}$ \cite{pegasus}; while the second inequality  holds with probability  at least $1-e^{-2Nt^2}$ using Hoeffding's inequality for  $\bar{\xi}_N := ({1}/{N}) \sum_{i=1}^N{\xi_i}$. $\hfill \blacksquare$

Note that under the i.i.d. assumption on $e_i := \|\bphi(\bx_i) - \hat{\bphi}(\bx_i)\|_\mathcal{H}^2$, random variables  $\xi_i$ are also i.i.d., rendering the conditions of Propositions 4 and 5 equivalent.

Next, we study the performance of linear SVMs trained on the set $\{\bz_i,y_i\}_{i=1}^N$, where $\bz_i := \bK_{\mathcal{S}}^{1/2} \bA \bq_i$; that is, the linear function $g(\bz) = \bw^\top \bz$ is learned by finding 
\begin{align}\label{primal_apprx_linear}
\bw^* =\arg\min_{\bw\in \mathbb{R}^{{r}}}&\; {\dfrac{1}{2} \|\bw\|^2 + 
	\dfrac{C}{N}\sum_{i=1}^N {\max \{0,1-y_i \bw^\top \bz_i\}}}.
\end{align}
The following result asserts that the classifiers learned through \eqref{primal_apprx}  and \eqref{primal_apprx_linear} can afford identical generalization capabilities. 

\noindent{\bf{Proposition 6}}.  \emph{The generalization capability of   classifiers  \eqref{primal_apprx} and \eqref{primal_apprx_linear} is identical, in the sense that $\hat{\bar{\bw}}^{*\top} \hat{\bphi}(\bx) ={\bw}^{*\top} \bz$.}

\noindent \emph{Proof:} Since for the low-rank approximation of the kernel matrix we have $\hat{\bK} = \bZ^\top \bZ$, then \eqref{dual_apprx} and \eqref{primal_apprx_linear} are equivalent, and consequently $\bw^* = \sum_{i=1}^N \hat\alpha_i^* y_i \bz_i$. Now, one can further expand $\hat{\bar{\bw}}^{*\top} \hat{\bphi}(\bx)$ and $\bw^{*\top} \bz$ to obtain
\begin{equation*}
\hat{\bar{\bw}}^{*\top} \hat{\bphi}(\bx) = \sum_{i=1}^N \hat{\alpha}_i^* y_i \hat{\bphi}^\top(\bx_i) \hat{\bphi}(\bx) =  \sum_{i=1}^N \hat{\alpha}_i^* y_i \bq_i^\top \bA^\top \hat{\bPhi}_\mathcal{S}^\top \hat{\bPhi}_\mathcal{S}\bA \bq
 \end{equation*}
 and $
\bw^{*\top} \bz = \sum_{i=1}^N \hat{\alpha}_i^* y_i \bz_i^\top \bz = \sum_{i=1}^N \hat{\alpha}_i^* y_i \bq_i^\top \bA^\top \hat{\bPhi}^\top_{\mathcal{S}} \hat{\bPhi}_{\mathcal{S}} \bA \bq$
where the equivalence  follows readily. $\hfill \blacksquare$

In addition to markedly reduced computational cost when utilizing  linear (L)SVM, 
our novel classifier can also be efficiently  trained online \cite{pegasus} as new data becomes available (or iteratively when the entire datasets can not be stored in memory which necessitates one-by-one acquisition). In this case, the proposed OK-FEB in Algorithm 4 can be run in parallel with the online classifier training, an attribute most suitable for big data applications. 
\vspace{-0.3cm}
\subsection{Stability analysis for kernel-based regression}

Consider now the kernel-based ridge regression task on the dataset $\{\bx_i,y_i\}_{i=1}^N$, namely 
\begin{equation}
\min_{\bbeta} \dfrac{1}{N}\|\by - \bK \bbeta\|_2^2+\lambda \bbeta^\top \bK \bbeta 
\end{equation}
which admits the closed-form solution $\bbeta^* = (\bK + \lambda N \mathbf{I})^{-1} \by $ \cite[~p.~251]{kernels}.
Alleviating the  $\mathcal{O}(N^2)$  memory requirement through low-rank approximation of  matrix $\bK$, the kernel-based ridge regression can be approximated by 
 \begin{equation}
\min_{\bbeta} \dfrac{1}{N}\|\by - \hat{\bK} \bbeta\|_2^2+\lambda \bbeta^\top \hat{\bK} \bbeta 
\end{equation}
whose solution is given as $\hat{\bbeta}^* = (\hat{\bK} + \lambda N \mathbf{I})^{-1} \by$. The following proposition bounds the mismatch between $\bbeta^*$ and $\hat{\bbeta}^*$.

\noindent{\bf{Proposition 7}}. \emph{If the random variables $e_i \in [0 \;, \; 1]$ are i.i.d., with mean $\bar{e}$, and $|y_i| \leq B_y$ for $i=1,2,...,N$, with  probability at least $1-2e ^{-2Nt^2}$, we have}
\begin{equation}
\|\bbeta^*-\hat{\bbeta}^*\|^2 \leq \dfrac{B_y}{\lambda^2} \sqrt{\bar{e}+t}(\sqrt{\bar{e}+t}+2) \;.
\end{equation}

\noindent \emph{Proof:} Following \cite{cortes}, we can write 
\begin{align*}
 \bbeta^*-\hat{\bbeta}^* &= (\bK + \lambda N \mathbf{I})^{-1} \by - (\hat{\bK} + \lambda N \mathbf{I})^{-1} \by \\ \nonumber 
 & = -\Big( (\hat{\bK} + \lambda N \mathbf{I})^{-1} (\bK-\hat{\bK}) ({\bK} + \lambda N \mathbf{I})^{-1} \Big) \by 
\end{align*}
where   we have used the identity $\hat{\bP}^{-1} - \bP^{-1} =  -  \bP^{-1} ( \hat{\bP} -  \bP) \hat{\bP}^{-1}$, which holds for any invertible matrices $\bP$ and $\hat{\bP}$. Taking the $\ell_2$-norm of both sides and using the Cauchy-Schwartz inequality, we  arrive at 
\begin{align}
 \|\bbeta^*-\hat{\bbeta}^* \|& \leq  \|(\bK + \lambda N \mathbf{I})^{-1} \|\|\bK-\hat{\bK}\| \|(\hat{\bK} + \lambda N \mathbf{I})^{-1}\|\|  \by \| \nonumber \\
 & \leq \dfrac{\|\bK-\hat{\bK}\| N B_y}{ \lambda_{\min}(\bK+\lambda N \mathbf{I}) \lambda_{\min}(\hat{\bK}+\lambda N \mathbf{I})} \nonumber \\
& \leq \dfrac{B_y \|\bK-\hat{\bK}\|_2}{\lambda^2 N} \;.
\end{align}

Using the inequality $\| \bP\|_2 \leq \|\bP\|_F$ along with Proposition 4,  yields the bound with probability $1-2e^{-2Nt^2}$. $\hfill \blacksquare$

\section{Numerical tests}\label{sec:tests}
This section presents numerical evaluation of various performance metrics to test our proposed algorithms using both synthetic and real datasets. In subsection 7.1, we empirically study  the proposed batch and online feature extraction algorithms using a toy synthetic dataset.  In subsection 7.2, we focus on the  tracking capability of the proposed OK-FEB and demonstrate its performance in terms of the evolution of average LS-fitting error obtained at iteration $n$ as  $({1}/{n}) \sum_{\nu=1}^n{\| \boldsymbol\phi(\bx_\nu) - \bar{\bL}[n] \bq_\nu \|_{\mathcal{H}}^2}$. Regarding the kernel matrix approximation performance, given a window size $N_{\text{wind}}$, we have ($N-N_\text{wind}$) windows  in a    dataset  of size $N$. Consequently,  the  mismatch of kernel matrix approximation is averaged over all such windows, and it is thus obtained as \[\dfrac{1}{N-N_\text{wind}}\sum_{w=1}^{N-N_{\text{wind}}} \Big(\dfrac{1}{N_{\text{wind}}} \|\mathbf{K}_w - \mathbf{\hat{K}}_w\|_F\Big)\]
 where $\mathbf{K}_w$ and $\hat{\mathbf{K}}_w$ are the kernel matrix and its approximation over the data vectors in the $w$-th window. Finally, in subsection 7.3 we test how well OK-FEB approximates the kernel-based classification and regression modules, and compare its performance with competing alternatives. 

\subsection{Kernel-based feature extraction: Batch vs. online}
Performance of Algorithms 1, 2 and 3 on solving the minimization \eqref{onlineSubspace} is tested using synthetically generated data arriving in streaming mode with  $\nu=1, 2, \dots, 5,000$. 
The  test involves generating  two equiprobable classes of $3 \times 1 $ data vectors $\{\bx_\nu\}$, each uniformly drawn from the surface of a sphere centered at the origin with radius $R_{c1} = 1$ or $R_{c2} =2$, depending on whether its label $y_\nu$ equals $1$ or $-1$, respectively. Noise drawn from the Gaussian distribution $\mathcal{N}(\mathbf{0}_{3\times 1},\sigma^2 \mathbf{I}_{3\times 3})$ is added to each $\bx_\nu$, with $\sigma^2$ controlling the overlap between the two classes.  Linear classifiers can not correctly classify  data generated in this manner. For this reason, the Gaussian kernel $\kappa(\bx_i,\bx_j) = \exp({-\|\bx_i-\bx_j\|_2^2}/{\gamma})$    was used with $\gamma=100$.  The online schemes can solve the problem on-the-fly, while the batch Algorithm 1 is also employed to solve  \eqref{onlineSubspace} offline. We compare the overall LS fit given by the subspace update $\bar{\bL}[n]$ using the three different solvers across time (iteration) index $n$.The parameters for the OK-FE solvers are chosen as $\mu_{n,L} \propto {1}/{n}$, $\mu_{n,A}\propto {1}/{n^2}$,  $\lambda=10^{-3}$,  and the maximum number of iterations in the batch solver is set to $I_{\max}=50$. 

Figure 1(a) depicts  how stochastic low-complexity updates of $\bA$ in the online solvers ensure convergence of the average LS cost to  the high-complexity  batch solution for $r=7$. When $n$ is small, the low-rank approximation is accurate and the resulting LS error in Batch-KFE is small. Note however that LS is nonzero for $n<r$, due to regularization. As $n$ increases, the number of vectors in the batch minimization also increases, while $r$ is fixed. Thus, the fitting problem becomes more challenging and the LS error increases slightly until $n$ is large enough and the $n$ data vectors are representative of the pdf from which data is drawn - a case that the LS fit stabilizes. Fig. 1(b) plots the convergence curve for Algs. 2 and 3.  
	
	  While the Gaussian kernel that was adopted here is the most widely used type, other kernels are also applicable (e.g. polynomial kernels).	Although it goes beyond the scope and claims of this paper, similar to all kernel-based schemes, the effect of not knowing the ideal kernel can be mitigated via data-driven multi-kernel approaches \cite{multikernel07,multikernel_bach}. Plotted in Fig. 2 is the fitting error for different kernels with different parameters  versus $r$ to highlight this issue (in Gaussian kernel, $\gamma=2\sigma^2$).   

 In addition, Fig. 3 plots the evolution of the average LS  cost across iterations for different choices of parameters $(r,B)$ in the OK-FEB solver. Note that relative to the batch Alg. 1 that incurs complexity $\mathcal{O}(N^2r)$ per iteration, OK-FE exhibits similar performance at much lower complexity  $\mathcal{O}(Nr^3+NDr)$.
 
\begin{figure}[t]
\label{fig:okfe_vs_batch}
	\begin{minipage}[b]{1\linewidth}
		\centering
		{\includegraphics[scale=0.5, center]{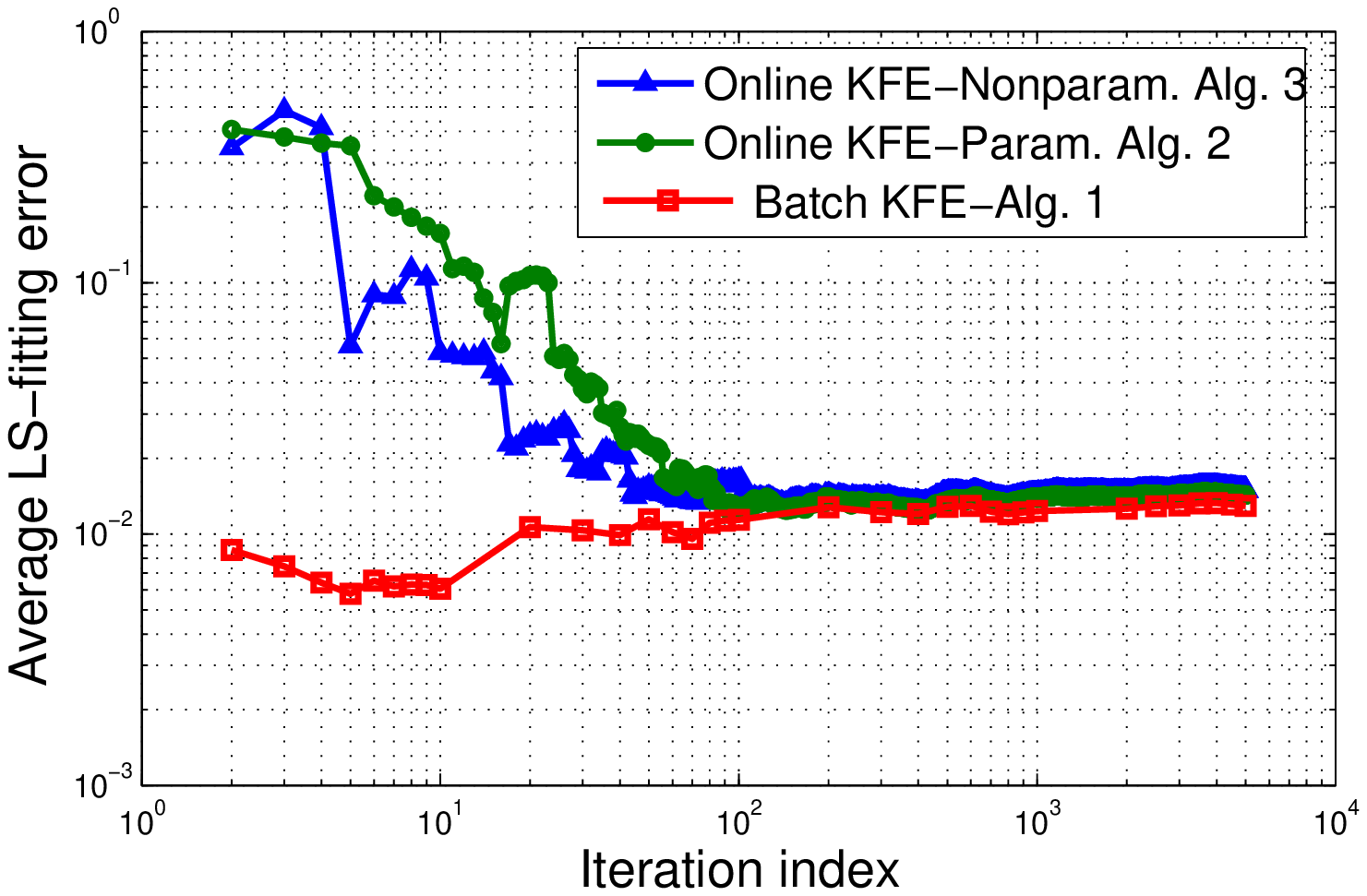}} %
		\vspace{-0.2cm}
		\centerline{(a)  }\medskip
	\end{minipage} 
	\begin{minipage}[b]{1\linewidth}
		\centering			
		{\includegraphics[scale=0.75, center]{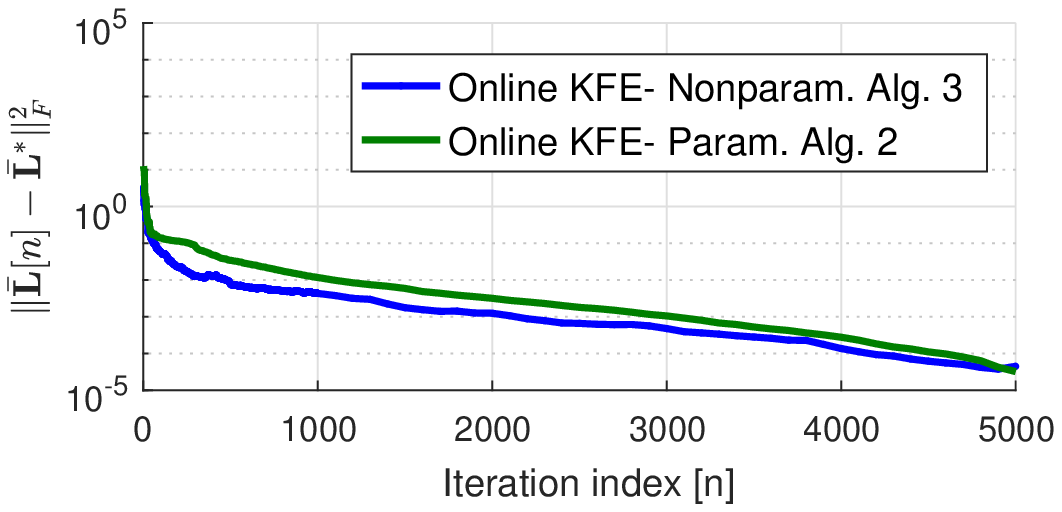}} %
		\vspace{-0.2cm}			
		\centerline{(b) }\medskip	
	\end{minipage}

\caption{ LS-fit versus iteration index for the synthetic dataset (a), and convergence curve for the subspace iterates (b).}
\end{figure}
%
\begin{figure}[ht]
\label{fig:kernels}
	\includegraphics[scale=0.6, center]{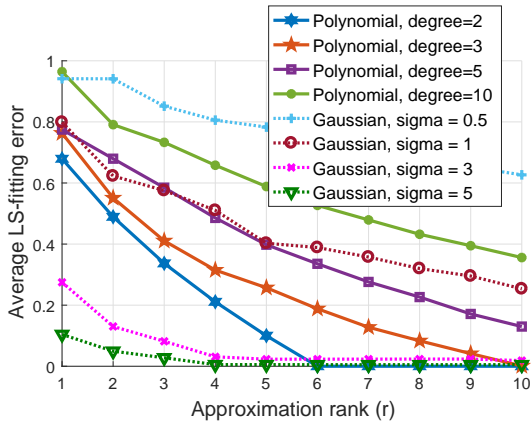} 
	\caption{  LS-fit of OKFE for different choices of polynomial and Gaussian kernels with different parameters}
	\vspace{-0.2cm}
\end{figure}

\begin{figure}[ht]
	\includegraphics[scale=0.45, center]{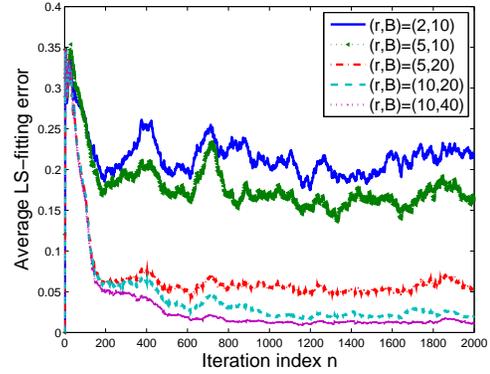} 
	\caption{  LS-fit for different choices of $(r,B)$ using OK-FEB}
	\vspace{-0.2cm}
\end{figure}

\subsection{Dynamic subspace tracking}
In this subsection, we assess efficiency of the novel approach  in tracking dynamic subspaces using synthetic and real-world datasets. 
\subsubsection{Synthetic data}
We generated a set of $N=2,000$ data vectors in $\mathbb{R}^3$. For $n=1,... , 1000$ the data  were drawn  from the surface of the sphere given by the manifold $(x_1/3)^2+x_2^2+x_3^2=1$, while for $n=1,001 , ... , 2,000$ they were sampled from the surface of the spheroid $x_1^2+(x_2/3)^2+(x_3)^2=1$, in Fig. 4. Plotted in Fig. 5 is the LS error of the low-rank feature extraction with $r=10$ and kernel parameter $\gamma=2$ (averaged by a window of length 200 for improved visualization) across time $n$. To enable tracking, the step size at sample index $\nu$ is chosen as $\mu_\nu = 1/\|\bq_\nu\|_2$. As the plot suggests, the change of the manifold at $n=1,000$ can be spotted by the rise in the LS error. The tracking capability of OK-FEB enables the subspace to adapt to the change in the underlying manifold. However, within a window of fixed subspace, namely for $1<n<1,000$, and $1,200<n<2,000$, and especially for small budget  $B=2r$, the budget maintenance policy in Alg. 3 outperforms the FIFO budget maintenance policy by carefully discarding the SVs whose exclusion least distorts the learned subspace.  Among the budgeted algorithms, setting small $\beta$ leads to a forceful exclusion of relatively older vectors, and thus adaptation to the new subspace at $t=1000$ takes place faster. In contrast, having small $\beta$ reduces the capability of fine tuning to the underlying subspace  when it is not changing. This is corroborated by the lower curve for $\beta=0.9$ versus $\beta=1$ during the subspace change, while $\beta=1$ gives lower error when subspace is not changing. Budget size  $B=2r$ demonstrates such effects more clearly as  smaller $B$ requires a more careful selection of the support vectors, hence emphasizing the effect of parameter $\beta$. The performance of the batch solver with no budget size constraint  is also plotted, whose average fitting error is worse than that of budget size $B=5r$ and is  similar to the very restrictive budget size $B=2r$. This is contributed to the fact that in the batch solver, the union of two subspaces is approximated by low-rank $r$, and thus the performance is inferior to the proposed online approach which is capable of tracking the underlying subspace. Overall, given the dynamics of a particular dataset, selection of $\beta$ directly sets the operation mode of our subspace learning, and  is tunable to the  pace of dynamics.

\begin{figure}[] 
\includegraphics[scale=0.4, center]{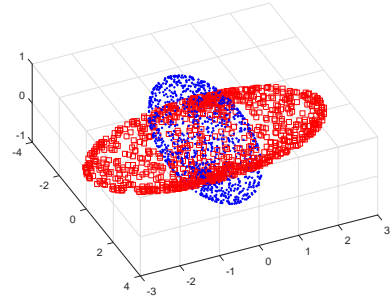} %
 
\caption{{Visualization of the nonlinear synthetic manifolds}} %
 
\end{figure}

\begin{figure}[] 
	\includegraphics[scale=0.6, center]{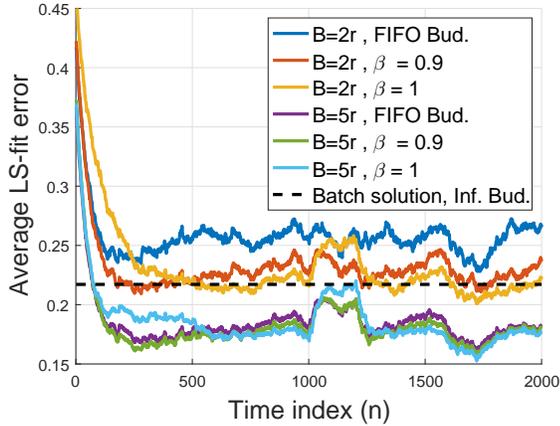} %
	\caption{{LS-fitting error of  dynamic   dataset versus time}} %
\end{figure}

Average mismatch of $\hat{\bK}$ found from  OK-FEB for various values of rank $r$ and choice of $B=2r$ is plotted in Fig. 6, and is compared with KPCA as well as  state-of-the-art variations of the Nystrom approximation, namely Improved Nystrom \cite{impnyst}, SS-Nystrom \cite{ssnyst}, and MEKA \cite{meka}.  Considering the dynamic nature of the data, the mismatch is evaluated over a moving window of length $N_{wind}=100$, and averaged over all such windows. As the plot suggests, OK-FEB  outperforms competing alternatives and  better suites  datasets with dynamic subspaces.
\begin{figure}
\includegraphics[scale=0.57, center]{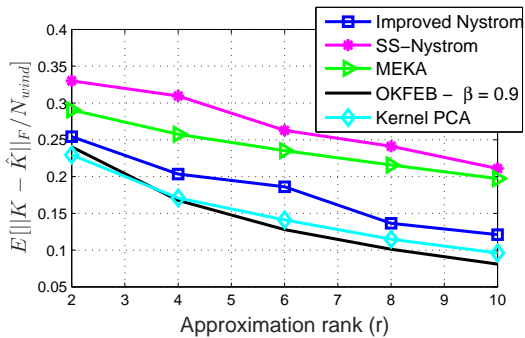} %
\caption{{Average kernel  mismatch of  dynamic   data }} %
\vspace{-0.5cm}
\end{figure}
\vspace{-0.3cm}

\subsection{Real-data on physical activity tracking}

In this subsection, we test performance of OK-FEB on the  physical activity monitoring dataset PAMAP2 \cite{pamap}.  The dataset contains $N=129,200$ measurements from 3 Colibri  wireless inertial measurement units (MU) worn by 9 subjects during different physical activities, such as walking and cycling. The MUs are placed on the dominant arm, chest, and dominant ankle of the subjects, each recording  13 quantities including  acceleration and  gyroscope data  with sampling frequency 100Hz. We discarded all  measurement vectors with missing entries, idle state measurements, and first and last $1,000$ measurements of each activity, as they correspond to transient states. The tests are performed on data corresponding to subject number 1, and can be similarly repeated for other subjects as well.

The data is  fed to OK-FEB  with $(r,B) = (10,15)$, and step size  set to $\mu_t=1/\|\bq_t\|_2$. LS error given by the nonlinear feature extraction (averaged over a window of length 200 for improved visualization) is plotted in Fig. 7 across time. Every activity is also coded to a number in $(0 , 1]$, and plotted in the same figure versus time to highlight the activity  changes over time. As the figure illustrates, different activities correspond to different manifolds,   each of which can be approximated with a certain accuracy via dynamic subspace learning and feature extraction. Introducing the forgetting factor $\beta<1$  enhances the learning capability. Table \ref{table:1} reports the average LS-error and its variance for different activities using various budget maintenance strategies, with $\beta=1, 0.9$, and the FIFO strategy.

%

\begin{figure}
\includegraphics[scale=0.55, center]{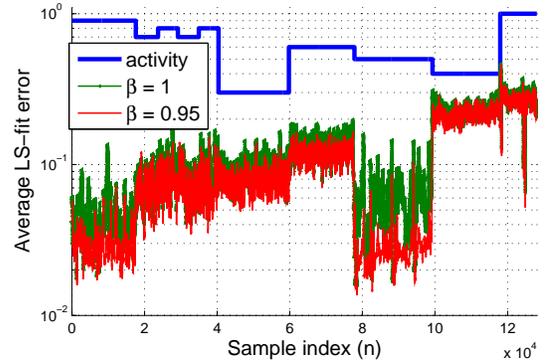}
\caption{{ LS-fitting error of the PAMAP2 dataset versus time }}
\end{figure}

\begin{table}[]
	\caption{ Mean and variance of LS-fitting error of the   extracted features with   $(r,B)=(10,15)$ for   different  activities using different budget maintenance  strategies  }
	\vspace{-0.2cm}
	\begin{center}
		\begin{tabular} {|c|c|c|c|c|c|c|c||c|c||c|c|c|}
			\hline
			Code & Activity & $\beta = 1$& $\beta = 0.9$& FIFO Bud.\\
			\hline 
			0.3 & Walking& $0.099$&$0.074 $&$0.074 $ \\ 
					 & & $ \pm  0.016$&$ \pm 0.012$&$ \pm 0.012$ \\ 
			\hline 
				0.4 & Running  &$ 0.227 $&$0.187$&$0.187 $	 \\
			 &   &$  \pm 0.025$&$ \pm 0.022$&$\pm 0.022$	 \\								\hline 
					0.5 & Cycling& $0.058 $&$0.028 $&$0.028$ \\
					 & & $ \pm 0.027$&$ \pm 0.012$&$ \pm 0.12$
					\\ \hline

					0.6 & Nordic & $0.130 $&$0.103 $&$0.103 $\\ 
					 &  Walking& $ \pm 0.020$&$ \pm 0.016$&$ \pm 0.016$\\ \hline
					0.7 & 	Ascending &$0.079 $&$0.063 $&$0.063$\\ 
					 & 	 Stairs&$ \pm 0.022$&$ \pm 0.018$&$\pm 0.018$\\ \hline					
					0.8 & Descending &$0.094 $&$0.066 $&$0.065 $\\ 
						&  Stairs&$\pm 0.021$&$ \pm 0.016$&$ \pm 0.016$\\ \hline
						0.9 & Vacuum & $0.045$  &$ 0.029$ & $0.029$\\ 
					 &  cleaning& $\pm0.013$  &$ \pm 0.008$ & $\pm 0.008$\\ \hline
					1.0 & Rope & $0.272 $&$0.238 $&$0.238 $\\
					&  jumping& $\pm 0.063$&$ \pm 0.057$&$ \pm 0.057$\\ \hline																									

		\end{tabular}\label{table:1}
	\end{center}

\end{table}

Similar to Fig. 6,  Fig. 8 depicts the average mismatch of kernel matrix approximation of OK-FEB with  $B=1.5r$ for the PAMAP2 dataset. Comparison with the competing  Nystrom variations in \cite{impnyst} and \cite{meka} clearly demonstrates the advantage of OK-FEB with forgetting factor $\beta=0.9$. Due to the large number of data vectors, KPCA and SS-Nystrom could not  be implemented.

\begin{figure}[t]
	\includegraphics[scale=0.6, center]{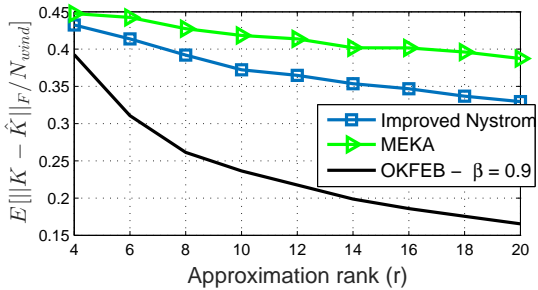}
	\caption{{Average kernel  mismatch of   PAMAP2 dataset  }}
	\vspace{-0.2cm}
\end{figure}

\vspace{-0.2cm}

\subsection{Online regression and classification} 
In this subsection, the generalization capability of the online linear classification and regression modules based on the features $\bZ$ returned by OK-FEB is tested.
We compare the  performance of    linear regression and classification as well as competing online kernel-based learners including (unbudgeted) Perceptron \cite{perceptron}, (unbudgeted) Norma \cite{kivinen}, (unbudgeted) online gradient descent (OGD) \cite{pegasus}, (unbudgeted) online dictionary learning (ODL),  and budgeted online gradient descent (BOGD) \cite{BSGD}, Forgetron \cite{forgetron}, Projectron \cite{projectron}, and budgeted passive-aggressive algorithm (BPA) \cite{BPA} with our novel OK-FEB, where the acquired features $\bz_n$ are fed to online linear Pegasus\cite{pegasus} and regularized-LMS solvers for the classification and regression tasks, respectively. 
The size and specifications of the dataset used are listed in Table \ref{table:2}, and are accessible from the LIBSVM website{\footnote {\url{http://www.csie.ntu.edu.tw/~cjlin/libsvmtools/} } or the UCI machine learning repository} \footnote{\url{http://www.ics.uci.edu/~mlearn/}}. The parameter values used per dataset are reported in Table \ref{table:2}.  In particular, tuning of the Frobenious norm regularization and kernel bandwidth parameters are done via cross validation over a discretized grid. Regarding the budget, to ensure stability of the algorithm it suffices that we set $B>r$, while it has been observed that setting $B$ very high yields only marginal improvement in terms of accuracy. Finally, for the selection of $r$, we test an increasing sequence of values starting from $r=2$ and gradually increasing until the  improvement in terms of fitting error becomes negligible. The aforementioned process is typically used to determine the minimum required complexity of parametric models (e.g., order-adaptive least-squares \cite{Kay_estimation_book}). 
	The censoring threshold $\epsilon$ is set using a moving-average of  LS-error values for the past $100$ data vectors. 

\begin{table}[t]
	\centering
	\caption{Specifications of datasets.}
	\begin{tabular}{c c c  c c c c c} 
		\hline
		dataset  & $D$&  $N$& $r$& $B$& $\gamma$& $C$ \\ [0.5ex] 
		\hline\hline
		Adult &123& 32K & $50$& $1.2r$& $20$&		 $10$ \\
		CADATA & 8&20.6K& $5$&$1,5r$&$7\times10^7$&$0.01$\\
		Slice& 384& 53.5K & $10$&$1.2r$&$50$&$0.01$\\
		Year & 90 & 463.7K& $10$ & $1.2r$&$5\times 10^7$&$0.01$ \\
		\hline
	\end{tabular}
	\label{table:2} 

\end{table}

Classification and regression accuracy as well as run time are plotted versus iteration index. Perceptron, Norma, ODL, and OGD are unbudgeted algorithms, and their SV sets (dictionary atoms in ODL) grow as iteration index increases in Fig. 9. Although the accuracy of these algorithms can serve as a benchmark, their run time grows the fastest. Thus, for the  ``Year'' dataset ($N \gg$), the mentioned algorithms are run only over $10\%$ of the data vectors. 
As these tests demonstrate,  among the budgeted algorithms, OK-FEB reliably approximates the kernel function through the extracted features, thus offering  more accurate classification and regression performance when compared to existing alternatives.
\begin{figure}[htb]
	
	\begin{minipage}[b]{0.49\linewidth}
		\centering
		{\includegraphics[width=0.99\textwidth,height=2.3in]{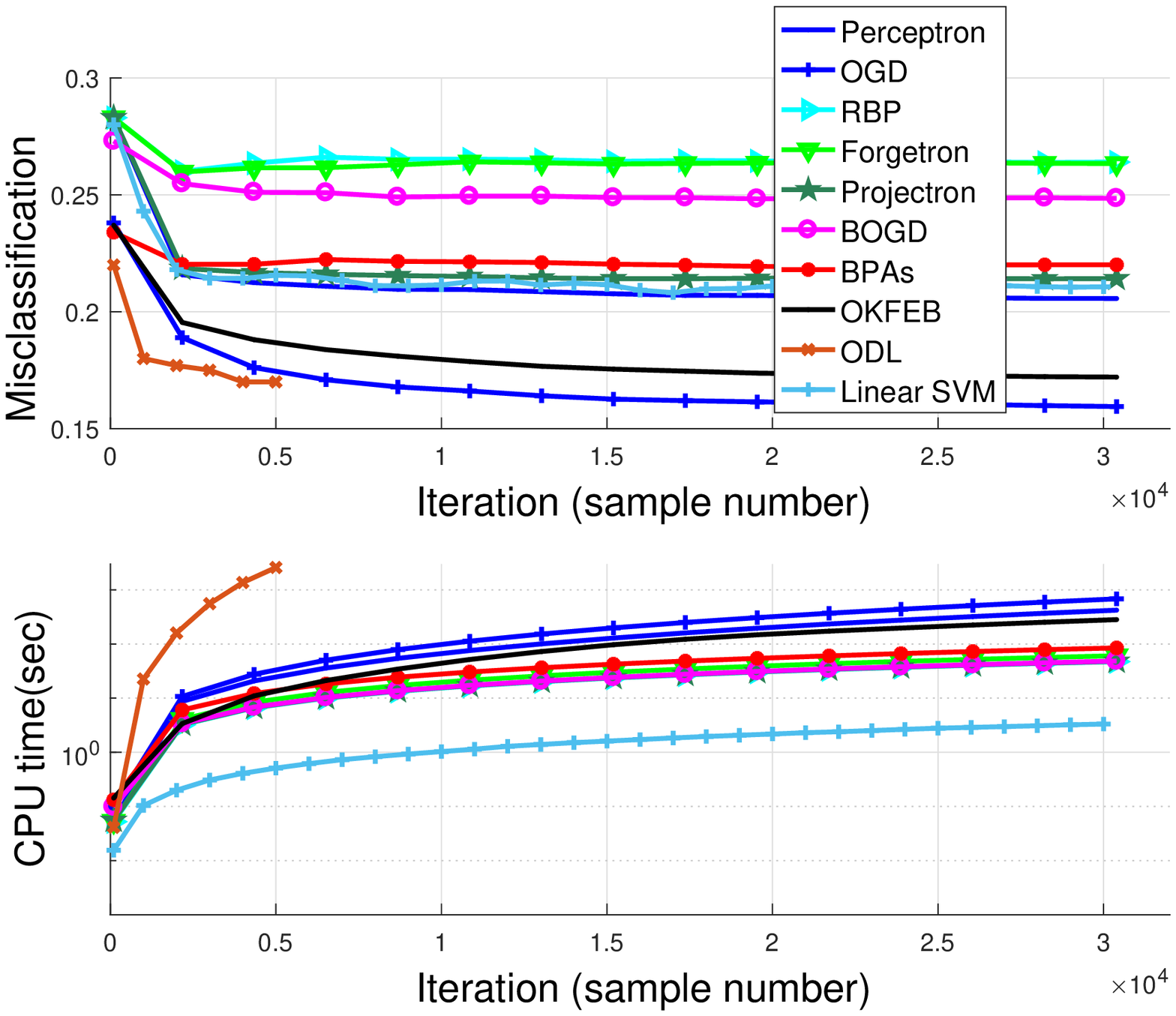}} %
 \vspace{-0.2cm}
	\centerline{(a)  }\medskip
	\end{minipage}
		\begin{minipage}[b]{0.49\linewidth}
			\centering			
			{\includegraphics[width=0.99\textwidth,height=2.3in]{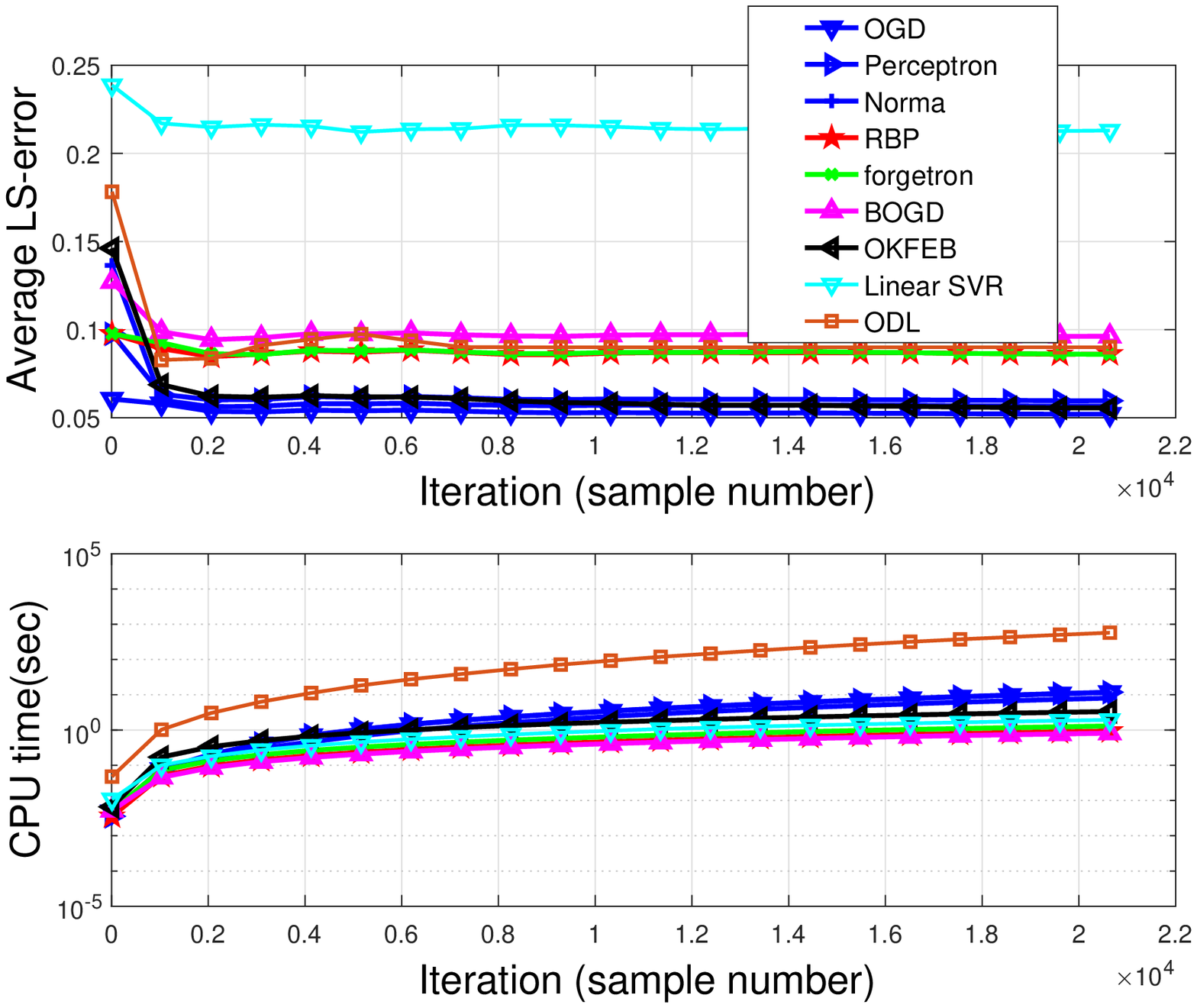}} %
             \vspace{-0.2cm}			
			\centerline{(b) }\medskip
		\end{minipage}
	

			%
%
%
		\begin{minipage}[b]{0.49\linewidth}
			{\includegraphics[width=0.99\textwidth,height=2.3in]{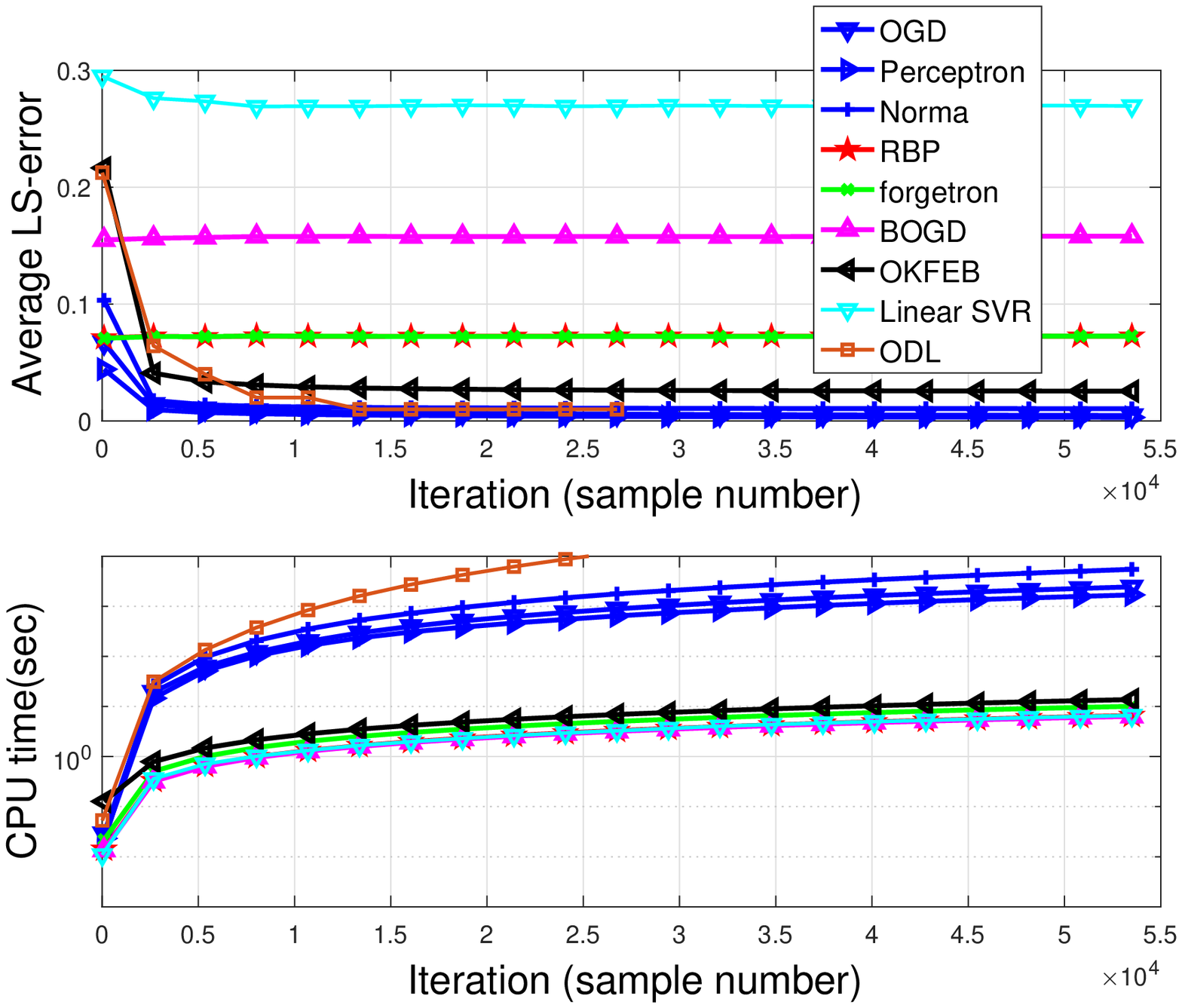}} %
			
	\centerline{(c)  }\medskip
		\end{minipage}
		\begin{minipage}[b]{0.49\linewidth}

			\centerline{\includegraphics[width=0.99\textwidth,height=2.3in]{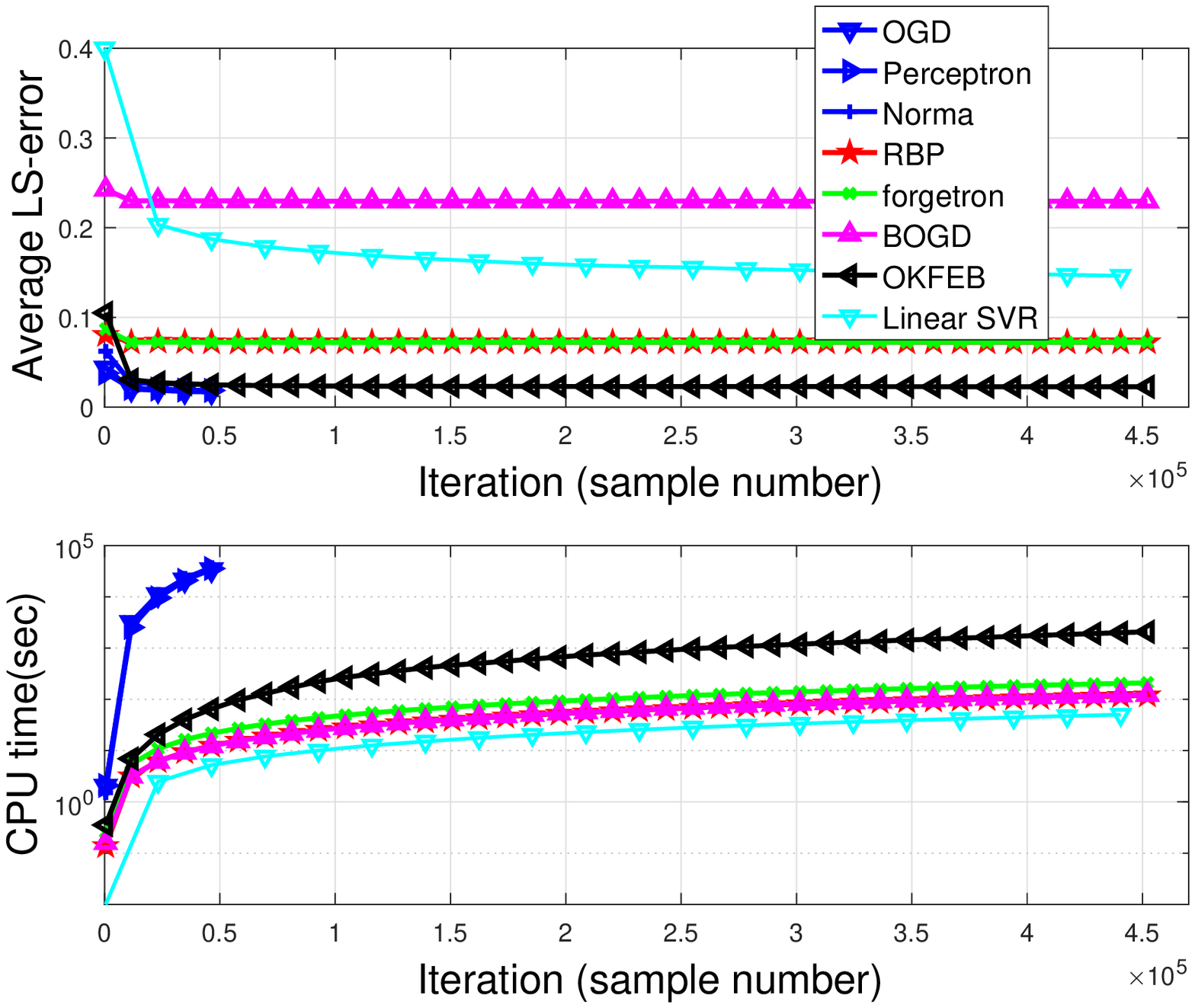}} %

		\centerline{(d) }\medskip
		\end{minipage} %
		\vspace{-0.5cm}
	\center{	\caption{{Online classification tests on (a) Adult,  and regression tests on (b) CADATA, (c) Slice, and (d) Year datasets.}}}
\vspace{-0.5cm}
\end{figure}
	
%
%
%
%
%
%
%

\section{Concluding Remarks}
{\em Low-complexity feature extraction} algorithms were introduced in this paper to markedly improve  performance of kernel-based learning methods  applied to large-scale datasets. The novel approach begins with a generative model having data  mapped to a high-(possibly infinite-) dimensional space, where they  lie close to a {\em linear} low-rank subspace, the tracking of which enables effective feature extraction on a budget. The extracted features can be used by fast linear classifiers or regression predictors at scale. 

 Offline and online solvers of the subspace learning task were developed, and their convergence was studied analytically. To keep the complexity and memory requirements within affordable levels, {\em budgeted} algorithms were devised, in which the number of stored data vectors is restricted to a  prescribed budget. Further analysis provided performance bounds on the quality of the resultant kernel matrix approximation, as well as the precision with which kernel-based classification and regression tasks can be approximated by leveraging budgeted online subspace-learning and feature-extraction tasks. 
 
 Finally, online subspace tracking and nonlinear feature extraction for dynamic datasets as well as  classification and regression tests on synthetic and real datasets  demonstrated the efficiency of OK-FEB with respect to competing alternatives, in terms of both accuracy and run time. 
\vspace{-0.2cm}

\section{Appendix}

\noindent 
\emph{Proof of Proposition 2}: The proof of the proposition is inspired by \cite{morteza_jstsp} and \cite{Mairal}, and is sketched along the following steps.

\noindent 
{\bf Step 1}. First, we judiciously introduce a surrogate for  $F_n(\bar\bL)$ whose minimizer coincides with the SGD updates in \eqref{LSGD}. 

To this end, we have that $\min_{\bq} {f_\nu} (\bx_\nu; \bar\bL, \bq ) \leq  {f_\nu} (\bx_\nu; \bar\bL, \bq[\nu] )$; hence, $\hat{F}_n(\bar\bL) := (1/n) \sum_{\nu=1}^n f_\nu(\bx_\nu;\bar\bL,\bq[\nu]) $ upper bounds the cost function, namely $F_n(\bar\bL)\leq\hat{F}_n(\bar\bL),\; \forall \bar{\bL}$. Further approximating $f_n$ through a second-order Taylor's expansion at the previous subspace update $\bar\bL[n-1]$, we arrive at
\begin{align}
\tilde{f}_n (\bx_n;\bar\bL,\bq[n])& =  f_n(\bx_n;\bar\bL[n-1],\bq[n]) \\&+\text{tr}\{ \nabla_{\bar\bL} f_n(\bx_n;\bar\bL[n-1],\bq[n]) (\bar\bL-\bar\bL[n-1])^\top\}\nonumber \\&+ \dfrac {\gamma_n}{2} \|\bar\bL-\bar\bL[n-1]\|_{HS}^2 \nonumber \;.
\end{align}
By choosing $\gamma_n \geq \|\nabla^2_{\bar\bL}f_n(\bx_n;\bar\bL[n-1],\bq_n)\|_{\mathcal{H}}= \|(\bq[n]\bq^\top[n]) \otimes \mathbf{I}_{\bar{D}} + (\lambda/n) \mathbf{I}_{r\bar{D}}\|_{\mathcal{H}}$ and using the norm properties in the Hilbert space,
the following can be verified: 

(i) $\tilde{f}_n$ is locally tight; i.e., $\tilde{f}_n(\bx_n;\bar\bL[n-1],\bq[n]) = {f}_n(\bx_n;\bar\bL[n-1],\bq[n])$;

(ii) gradient of $\tilde{f}_n$
is locally tight; i.e., $\nabla_{\bar\bL}\tilde{f}_n(\bx_n;\bar\bL[n-1],\bq[n]) = \nabla_{\bar\bL}{f}_n(\bx_n;\bar\bL[n-1],\bq[n])$;
and 

(iii) $\tilde{f}_n$ {\it{globally}} majorizes the original instantaneous cost $f_n$; that is, $f_n(\bx_n;\bar\bL,\bq[n]) \leq \tilde{f}_n(\bx_n;\bar\bL,\bq[n])$, $\forall\, \bar\bL$.

Selecting now the target surrogate cost as 
$\tilde{F}_n(\bar\bL) =  \dfrac{1}{n}\sum_{\nu=1}^n \tilde{f}_\nu(\bx_\nu;\bar\bL,\bq[\nu])$ 
we have $F_n(\bar\bL)\leq \hat{F}_n(\bar\bL) \leq \tilde{F}_n(\bar\bL) , \forall \, \bar\bL$. 
Minimizing the cost $\tilde{F}_n (\bar\bL)$ amounts to nullifying the gradient, i.e., $\nabla_{\bar{\bL}}\tilde{F}_n(\bar{\bL}[n])=\mathbf{0}$, which yields  \cite{morteza_jstsp}
$
\bar\bL[n] = \bar\bL[n-1] - {\bar\gamma_n}^{-1} \bar\bG_n
$, with $\bar{\gamma}_n := \sum_{\nu=1}^n \gamma_\nu$. By setting  $\mu_n= 1/\bar\gamma_n$, the SGD-based update of  $\bar\bL[n]$ now coincides with the minimizer of $\tilde{F}_n (\bar\bL)$; that is, $\bar\bL[n] = \arg\min_{\bar\bL} \tilde{F}_n(\bar\bL)$.   

\noindent 
{\bf Step 2}. The second step establishes that the surrogate costs $\{\tilde{F}_n (\bar\bL)\}$ form a quasi-martingale sequence \cite{quasi}, and using tightness of the surrogate cost we deduce that $\lim_{n \rightarrow \infty} (F_n(\bar\bL[n])-\tilde{F}_n(\bar\bL[n]))=0$. Thus, the surrogate cost asymptotically converges to the original cost $F_n(\bar\bL)$.

\noindent 
{\bf Step 3}. Leveraging the regularity of $\bar{\cal L}(\bx_\nu; \bar\bL,\bq_\nu)$,  convergence of the cost sequence implies convergence of $\{\|\nabla_{\bar\bL}F_n(\bar\bL[n]) - \nabla_{\bar\bL} \tilde{F}_n(\bar\bL[n])\|_{\cal{HS}}\}$ to zero, which along with $\nabla_{\bar{\bL}}\tilde{F}_n(\bar{\bL}[n])=\mathbf{0}$, yields $\{\|\nabla_{\bar\bL}F_n(\bar\bL[n])\|_{\cal{HS}}\} \rightarrow \mathbf{0}$.  $\hfill\blacksquare$

{
}
\end{document}